\documentclass{article}

\usepackage{arxiv}

\usepackage[utf8]{inputenc} % allow utf-8 input
\usepackage[T1]{fontenc}    % use 8-bit T1 fonts
\usepackage{hyperref}       % hyperlinks
\usepackage{url}            % simple URL typesetting
\usepackage{booktabs}       % professional-quality tables
\usepackage{amsfonts}       % blackboard math symbols
\usepackage{nicefrac}       % compact symbols for 1/2, etc.
\usepackage{microtype}      % microtypography
\usepackage{lipsum}		% Can be removed after putting your text content
\usepackage{graphicx}
\usepackage{natbib}
\usepackage{doi}

\usepackage{ mathrsfs }
\usepackage{placeins}

\title{Integrating recurrent neural networks with data assimilation for scalable data-driven state estimation}

\date{September 24, 2021}	% Here you can change the date presented in the paper title
\author{\href{https://orcid.org/0000-0002-5223-8307}{\includegraphics[scale=0.06]{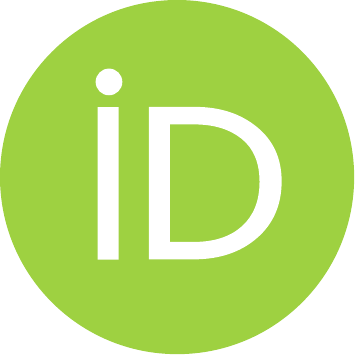}}\hspace{1mm} {Stephen G. Penny}\thanks{Corresponding author: Stephen G. Penny, Steve.Penny@noaa.gov},  \hspace{1mm} {Timothy A. Smith}, \hspace{1mm} {Tse-Chun Chen}, \hspace{1mm} {Hsin-Yi Lin}, \hspace{1mm} \href{https://orcid.org/0000-0002-6841-1661}{\includegraphics[scale=0.06]{Images_r1/orcid.pdf}}\hspace{1mm} {Michael Goodliff} \\
	Cooperative Institute for Research in Environmental Sciences\\
	University of Colorado Boulder\\
	Boulder, CO 80309 \\
	Physical Sciences Laboratory \\
	National Oceanic and Atmospheric Administration \\
	Boulder, CO 80305
	\And
	{Jason A. Platt} \\
	Department of Physics\\
	University of California San Diego \\
	La Jolla, CA 92093 \\
	\And
	\href{https://orcid.org/0000-0002-4690-6081}{\includegraphics[scale=0.06]{Images_r1/orcid.pdf}}\hspace{1mm} {Henry D.I. Abarbanel} \\
	Department of Physics\\
	University of California San Diego \\
	La Jolla, CA 92093 \\
	Scripps Institution of Oceanography \\
	La Jolla, CA 92037 \\
}

\renewcommand{\shorttitle}{Integrating Recurrent Neural Networks with Data Assimilation}

\hypersetup{
pdftitle={RNN-DA},
pdfsubject={q-bio.NC, q-bio.QM},
pdfauthor={Stephen G. Penny},
pdfkeywords={Data Assimilation, Recurrent Neural Networks, Reservoir Computing, Numerical Weather Prediction},
}

\begin{document}
\maketitle

\begin{abstract}
Data assimilation (DA) is integrated with machine learning in order to perform entirely data-driven online state estimation. To achieve this, recurrent neural networks (RNNs) are implemented as surrogate models to replace key components of the DA cycle in numerical weather prediction (NWP), including the conventional numerical forecast model, the forecast error covariance matrix, and the tangent linear and adjoint models. It is shown how these RNNs can be initialized using DA methods to directly update the hidden/reservoir state with observations of the target system. The results indicate that these techniques can be applied to estimate the state of a system for the repeated initialization of short-term forecasts, even in the absence of a traditional numerical forecast model. Further, it is demonstrated how these integrated RNN-DA methods can scale to higher dimensions by applying domain localization and parallelization, providing a path for practical applications in NWP.
\end{abstract}

% keywords can be removed
\keywords{Data Assimilation \and Recurrent Neural Networks \and Reservoir Computing \and Numerical Weather Prediction}

\section{Introduction}
Numerical weather prediction (NWP) requires two primary components: a computational forecast model and an initialization method, both of which have been demonstrated to contribute approximately equally to the steady improvement in forecast skill over the past 40 years. We seek to replace the computational forecast model with a data-driven surrogate model and integrate these two critical components. Weather forecast models typically push the boundaries of computational feasibility, even on the largest supercomputers in the world, with a drive towards increased grid resolutions and better resolved physical processes. The most sophisticated initialization methods require executing the forecast model many times, using iterative loops and ensembles of forecasts initialized from perturbed initial conditions. This creates a competing paradigm where computational resources must be balanced between model fidelity and initialization accuracy. As a result, the models serve as a major limiting factor in the development of new initialization methods.

The application of artificial intelligence and machine learning (AI/ML) methods in weather and climate is a rapidly growing activity. \cite{boukabara2021} described multiple instances in which AI/ML is being developed for target applications in operational NWP. An area of interest noted by \cite{boukabara2021} is the synergy between AI/ML and data assimilation (DA). \cite{abarbanel2018} noted deep connections between ML and DA, and in special cases mathematical equivalences (further details will be provided in an upcoming work \citep{abarbanel2021}). Recent approaches that have been applied to combine ML with DA include the application of a neural network design combined with a DA operation to train the network on noisy data \citep{brajard2020combining}, the application of artificial neural networks for correcting errors in numerical model forecasts in the DA cycle \citep{bonavita2020machine}, the use of a convolution neural network (CNN) to enforce conservation of mass in a DA procedure \citep{ruckstuhl2021}, the development of a NN-based tangent linear and adjoint model for use in variational DA methods \citep{hatfield_building_2021}, and an end-to-end application of a combined DA and model bias correction \citep{arcucci2021deep}.

Here we will focus on a simplified form of recurrent neural network (RNN), based on the reservoir computing (RC) paradigm, that can be used to replace the numerical model in the DA process. Integrating a data-driven model with DA techniques requires accurate characterization of dynamical error growth. We will demonstrate that the RNN architecture can produce sufficiently accurate representations of such error growth to the degree that the RNN-based models can replace key components of foundational DA algorithms such as the ensemble Kalman filter (EnKF) \citep{evensen1994} and the 4D variational method (4D-Var) \citep{talagrand1987, courtier1987, courtier1994}. Such key components include forecast error covariance statistics derived from ensemble forecasts, and the tangent linear model and its adjoint. 

\begin{figure}
    \centering
    \includegraphics[width=0.9\textwidth]{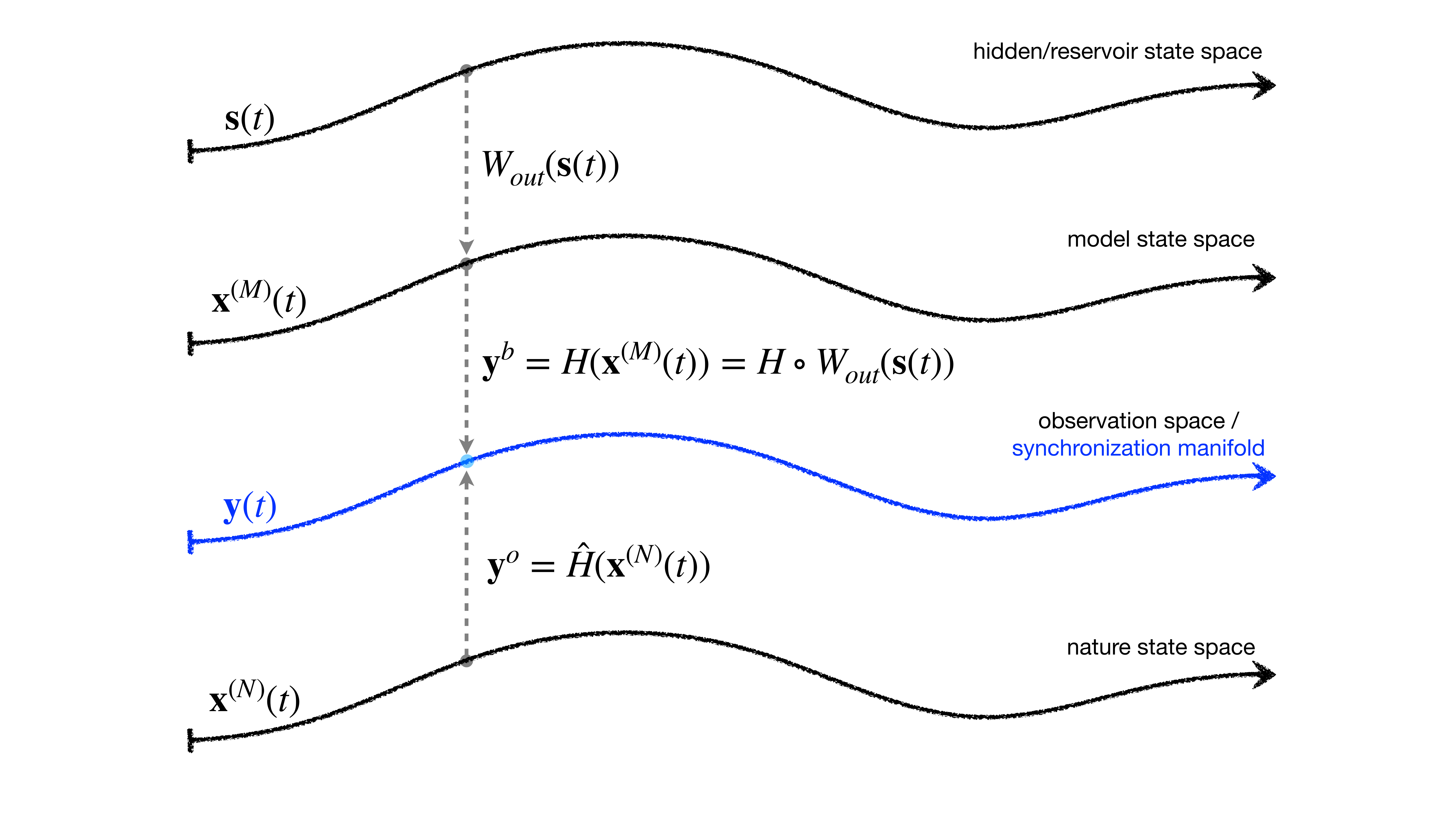}
    \caption{Data assimilation is applied to update the hidden/reservoir state space $\mathbf{s}(t)$. Observations $\mathbf{y}^o$ are sampled from the nature state space $\mathbf{x}^{(N)}(t)$, while the composition $H \circ W_{out}$ is used as an observation operator to map the hidden/reservoir state space to an equivalent representation that can be used to form the innovations $\mathbf{d}(t) = \mathbf{y}^o(t) - H \circ W_{out}(\mathbf{s}(t))$ in the observation space.}
    \label{fig:RNN-DA-diagram}
\end{figure}

We introduce a method to achieve this in a direct manner by applying DA to update the ‘hidden’ or ‘reservoir’ space of the RNN/RC dynamics (Figure \ref{fig:RNN-DA-diagram}). Results are shown assimilating both fully observed and sparsely observed dynamics, with a range of observational noise levels, using an RNN-based ensemble transform Kalman filter (ETKF) \citep{bishop2001adaptive, hunt2007efficient} and an RNN-based strong constraint incremental 4D-Var \citep{courtier1994}.

\section{Methods}
% Here is text on Materials and Methods.
%

The weather and climate communities regularly produce three primary categories of data: (1) observations of the Earth system, either drawn directly from in situ measurements or from satellite-based sensors, (2) simulations produced on regular gridded fields, generated from numerical models derived from basic principles of physics, and (3) retrospective analyses (reanalyses), that attempt to optimally combine the previous two categories to produce historical reconstructions of the atmosphere or other components of the Earth system. All of these datasets have inherent weaknesses. Observations are generally sparse and irregular, and can contain unpredictable errors. Numerical models have systematic errors that result from abstractions, approximations, and unresolved processes. Reanalysis products attempt to mitigate these weaknesses, but still inherit them to some degree. However, as these datasets have already been produced and archived, they are valuable resources that can be leveraged to develop data-driven methods.

We assume that for a realistic application either a long numerical model simulation or retrospective historical analysis is available as training data. Thus, we train the RNN to generate accurate predictions using a dedicated training dataset that resolves all components of the target dynamics. For this study, we use the output from the model described in section \ref{sec:source_data} to train our RNN models. We focus our attention on developing the capabilities to integrate the data-driven RNN model with an online DA process that repeatedly ingests new noisy and sparse observations, updates the state estimate of the system, and makes new short-term forecasts. This has applications ranging from operational NWP to the efficient reconstruction of historical Earth system states.

\subsection{Network Design}

Reservoir computing is a category of machine learning methods. It originated in the works of \cite{jaeger2001echo}, who introduced Echo State Networks (ESN), and \cite{maass2002} who introduced Liquid State Machines (LSM). Both methods assume that an input signal can be mapped to a fixed nonlinear system of higher dimension than the input. Such systems can be trained using a readout layer to map the state of the reservoir to the desired output. The result is a simple model that can reproduce the potentially complex dynamics of the original system. A survey is provided by \cite{lukosevicius2009}, while \cite{konkoli2017} provides further discussion on the generality of RC. Successful applications of RC have been demonstrated for the prediction of spatiotemporally chaotic dynamics \citep{chattopadhyay2020data, pathak2018model, platt2021forecasting}, and in particular of geophysical fluid dynamics \citep{arcomano2020machine, lin2021}. 

RNNs have long been used as a preferred ML method for cases in which temporal considerations are necessary. We use a basic RNN \citep{elman1990finding} with a simplified structure to produce a variant of the RC network. The RNN/RC design is useful for prediction because the ‘hidden’ or ‘reservoir’ state $\mathbf{s}(t_i)$ provides a short-term memory of the target system trajectory $\mathbf{x}(t)$ up to time $t_i$. Assuming an accurate forecast can be made of the hidden/reservoir state $\mathbf{s}(t_{i+1})$ by a well-trained RNN, then that forecast can be mapped to the target system space to produce a forecast $\mathbf{x}'(t_{i+1})$ of the true system state $\mathbf{x}(t_{i+1})$. 

As highlighted by \cite{Schrauwen07}, \cite{Steil2004} showed that the state-of-the-art learning rule for RNNs at the time had the same weight dynamics as the methods proposed by \cite{jaeger2001echo} and \cite{maass2002}. The \cite{Atiya2000} recurrent learning rule trains the output weights while the internal weights are only globally scaled up or down. Similarly, we classify the RNN model parameters as either `macro-scale' or `micro-scale'.  We note that what we call `macro-scale' parameters were called `global parameter' by \cite{lukosevicius2012}. In what follows, all matrix elements are classified as micro-scale parameters while all scalars are classified as macro-scale parameters. The general form of our RNN is given as,
 \begin{equation}\label{eqn:RC}
\mathbf{s}(t_{i+1}) = F(\mathbf{s}(t_i),\mathbf{x}(t_i))= l*f(\rho \mathbf{W}_{res}\mathbf{s}(t_i) + \sigma \mathbf{W}_{in}\mathbf{x}(t_i)) + (1-l)*\mathbf{s}(t_i),
 \end{equation}
 
 \begin{equation}\label{eqn:RC2}
\mathbf{x}'(t_{i+1})=G(\mathbf{s}(t_{i+1}))=\mathbf{W}_{out}(\mathbf{s}(t_{i+1})),
 \end{equation}
where $\mathbf{x}(t_i)$ is the system state at time $t_i$, provided from data, $\mathbf{s}(t_i)$ is the hidden/reservoir state, and $\mathbf{x}'(t_{i+1})$ is the predicted system state at the next time $t_{i+1}$. The parameter $\rho$ determines the spectral radius of the reservoir adjacency matrix $\mathbf{W}_{res}$, $\sigma$ scales the input signal mapped to the reservoir space by $\mathbf{W}_{in}$, and $l$ is the ‘leak’ parameter that gates new information into the system. We will use $f=\tanh$.

After training the RNN model parameters, we expect the predicted state $\mathbf{x}'(t_{i+1})$ to be close to the true system state $\mathbf{x}(t_{i+1})$. In a typical RNN, all parameters of the system described by equations (\ref{eqn:RC}) and (\ref{eqn:RC2}) are trained, usually by a gradient descent type optimization method. For RC, all model parameters in equation (\ref{eqn:RC}) are assumed fixed. Equation (\ref{eqn:RC}) is then iterated to generate a time series $\mathbf{s}(t)$ of hidden/reservoir states that corresponds to the training data $\mathbf{x}(t)$. The `readout' operator $\mathbf{W}_{out}$ in equation (\ref{eqn:RC2}) is typically assumed to be linear.

The micro-scale parameters in the $\mathbf{W}_{out}$ are trained (while the other parameters of our RNN remain fixed) by solving a regularized least squares equation using a loss function that targets forecasts that are one time step in the future \citep{jaeger2001echo}. The corresponding loss function is,
\begin{equation}
\mathscr{L}_{micro}\left(\mathbf{W}_{out}\right)=||\mathbf{W}_{out} \mathbf{S}_{data} - \mathbf{X}_{data}||^2 + \beta ||\mathbf{W}_{out}||^2,
\end{equation}

where $\mathbf{S}_{data}$ and $\mathbf{X}_{data}$ are matrices that comprise the vector-valued states $\mathbf{s}(t)$ and $\mathbf{x}(t)$ for the entire training dataset, ordered columnwise, and $\beta$ is a Tikhonov regularization parameter. 

To move closer towards the approach of the general RNN, in which all parameters are trained, we regard the scalars $l$, $\rho$, $\sigma$, and $\beta$ as macro-scale parameters subject to training. The matrix $\mathbf{W}_{res}$ is initialized with a spectral radius of 1 prior to training, but otherwise the values are assigned randomly using a uniform distribution centered at 0. For computational efficiency,  $\mathbf{W}_{res}$ is assumed to be sparse (i.e. only 1\% of entries are nonzero). The matrix $\mathbf{W}_{in}$ is initialized using a uniform random distribution with values ranging from -1 to 1.

An extended forecast $\mathbf{x}^f(t)$ is made with the RNN from time $t_0$ to $t_i$, for $i>0$, by recursively replacing the input state with the RNN prediction initialized from the previous time. Defined inductively, commencing with $\mathbf{x}(t_{0})$,

\begin{equation}\label{eqn:RCfcst1}
    \mathbf{x}^f(t_1) = \mathbf{W}_{out} \circ F(\mathbf{s}(t_{0}),\mathbf{x}(t_{0})),
\end{equation}
\begin{equation}\label{eqn:RCfcst2}
    \mathbf{x}^f(t_{i}) = \mathbf{W}_{out} \circ F(\mathbf{s}(t_{i-1}),\mathbf{x}^f(t_{i-1})).
\end{equation}
We note that due to the nature of chaotic dynamical systems (with leading Lyapunov exponent greater than 0), any error in one step of this recursion will accumulate and lead to exponential error growth over time.

The macro-scale parameters are trained using a nonlinear Bayesian optimization method, which uses a surrogate model representation of the loss function \citep{jones1998efficient, ginsbourger2010kriging}. For this macro-scale optimization we use a loss function that targets longer-range prediction,

\begin{equation}
\mathscr{L}_{macro}\left(\mathbf{x}^f(t)\right) = \sum_{i=1}^{M} \sum_{t=t_i}^{t_{i+N}} ||\mathbf{x}^{f}(t) -  \mathbf{x}(t)||^2 \exp{\left(-\frac{t-t_i}{t_{i+N}-t_i}\right)},
\label{eq:macro_loss}
\end{equation}

where $M$ represents the number of separate initial times $[t_1,t_2,t_3,...,t_M]$ used to make independent forecasts, selected randomly without replacement from the training dataset, and $N$ represents the number of time steps used for each forecast. We apply an exponential scaling term in order to account for the exponential growth of errors that is typical of chaotic dynamics. This term gives the forecast errors in the earlier portion of the forecast more weight, as this period is more relevant for cycled DA applications. 

Recall that by definition a hyperparameter is any design decision that is set before the learning process begins, is generally tunable, and can directly affect how well a model trains. The hyperparameters of this optimization are provided in the appendix in Table \ref{tab:hyperparams}. We note that \citet{griffith2019forecasting} similarly applied long forecasts at multiple initial times in a reservoir computing application to identify model parameters that resulted in the reconstruction of the full system attractor.

By computing the Jacobian of the forecast model defined by equations (\ref{eqn:RC}), (\ref{eqn:RC2}), (\ref{eqn:RCfcst1}), and (\ref{eqn:RCfcst2}) with respect to the hidden/reservoir state, we can determine the linear propagator $\mathbf{M}$ of the reservoir dynamics from time $t_i$ to $t_{i+1}$ as,
 \begin{equation}
\mathbf{W}=\rho\mathbf{W}_{res}+\sigma\mathbf{W}_{in}\mathbf{W}_{out},
 \end{equation}
 \begin{equation} \label{eqn:RNN-TLM}
\mathbf{M}_{[t_{i+1},t_i]}(\mathbf{s}(t_i))=\frac{dF}{d\mathbf{s}}=l*diag(1-\tanh(\mathbf{W}\mathbf{s}(t_i))^2 )\mathbf{W}+(1-l)*\mathbf{I}.
 \end{equation} 
The linear propagator describes the evolution of small perturbations from a reference trajectory. It can be used in DA, in particular 4D-Var, where it is called the tangent linear model (TLM). Further, the Lyapunov exponents of the system are determined by integrating the linear propagator from time $t=0\to\infty$ and computing the eigenvalues of the resulting system. Practical algorithms based on QR decompositions are provided by \citet{geist1990comparison}. For computational efficiency, we implement the TLM and its adjoint as linear operators to avoid matrix multiplications and allow for efficient matrix-vector operations applied within the iterative minimization schemes.

\subsection{Data Assimilation}
\label{sec:da_methods}

\citet{trevisan2010four}, \citet{trevisan2011}, and \citet{palatella2013lyapunov} showed that the number of observations needed to constrain any DA system is related to the number of non-negative Lyapunov exponents in the system. \citet{platt2021forecasting} indicated that reproducing the Lyapunov spectrum is critical to generating accurate predictions with reservoir computing models - with deviations from the true spectrum leading to significantly degraded forecast skill. Considering these points, we presume that even if the hidden/reservoir state space is large, if the RNN is trained to be sufficiently accurate such that the true Lyapunov spectrum is well approximated, then the number of observations required to constrain the hidden state space dynamics should be the same as is required to constrain the original system dynamics.

Following this presumption, we apply DA in the hidden/reservoir space of the RNN system, and apply the composition of an observation operator with the readout operator in order to compare hidden/reservoir states with observations of the original system. To test this approach, we apply two well known DA algorithms integrated with the RNN forecast model - the ensemble transform Kalman filter \citep{bishop2001adaptive, hunt2007efficient} and the strong constraint incremental 4-dimensional variational method (4D-Var) \citep{courtier1994}. 

From the perspective of operational forecasting, the RNN provides a simple and low-cost replacement for the production of essential information needed for the online DA cycle, such as forecast error covariance statistics and the tangent linear and adjoint model dynamics. From the machine learning perspective, the DA algorithms allow the RNN hidden/reservoir dynamics to be driven with a noisy and sparsely observed signal. We will show that in cases where the direct insertion of observations quickly corrupts the hidden/reservoir state and leads to inaccurate forecasts, the DA methods can produce valid reconstructions of the system state as well as viable initial conditions for short-term forecasts.

\cite{kalnay2006ensemble} described the ideal initial ensemble perturbations as those that effectively span the space defined by the analysis error covariance. We use ensemble forecast statistics produced by the RNN model to generate dynamically estimated forecast error covariance statistics, and then apply an ETKF to assimilate noisy observations of the true system state and estimate the analysis error covariance. \cite{bocquet2017} showed that the minimum ensemble size required to constrain a (non-localized) deterministic ensemble filter such as the ETKF is equal to the number of non-negative Lyapunov exponents of the system dynamics. Thus, we expect that with a well-trained RNN that closely approximates the correct Lyapunov spectrum, the minimum number of ensemble members needed to constrain the ETKF will be the same as the number of members needed to constrain the original system.

We define a new modified observation operator by composing the conventional observation operator $H()$, which maps from the system space to the observation space, with the readout operator $W_{out}()$, which maps from the hidden/reservoir space to the system space (see Figure \ref{fig:RNN-DA-diagram}). Our implementation of the ETKF follows the formulation of \citet{hunt2007efficient}. Let $\bar{\mathbf{y}}^b=H(W_{out}(\bar{\mathbf{s}}^b))$, where $\bar{\mathbf{s}}^b$ is the background ensemble mean hidden/reservoir state, and $\mathbf{Y}^b=H(W_{out}(\mathbf{S}^b))$, where the columns of $\mathbf{S}^b$ are ensemble perturbations around the mean, then

\begin{equation}
\tilde{\mathbf{P}}^a = \left[ \frac{k-1}{\gamma} \mathbf{I} + (\mathbf{Y}^b)^T \mathbf{R}^{-1}\mathbf{Y}^b\right]^{-1},
\end{equation}

\begin{equation}
\mathbf{W}^a=\left[(k-1)\tilde{\mathbf{P}}^a\right]^{\frac{1}{2}},
\end{equation}

\begin{equation}
\mathbf{S}^a=\mathbf{S}^b\mathbf{W}^a,
\end{equation}

\begin{equation}
\bar{\mathbf{w}}^a=\tilde{\mathbf{P}}^a (\mathbf{Y}^b)^T \mathbf{R}^{-1} \left(\mathbf{y}^o - \bar{\mathbf{y}}^b \right),
\end{equation}

\begin{equation}
\bar{\mathbf{s}}^a = \mathbf{S}^b\bar{\mathbf{w}}^a + \bar{\mathbf{s}}^b ,
\end{equation}

where $\mathbf{R}$ is the observation error covariance matrix, $k$ is the ensemble dimension, $\gamma$ is a multiplicative inflation factor, $\tilde{\mathbf{P}}^a$ is the analysis error covariance matrix represented in the ensemble perturbation subspace, $\mathbf{W}^a$ is applied as a transform operator to map the background ensemble perturbations to analysis ensemble perturbations, and $\bar{\mathbf{w}}^a$ determines the weighting coefficients of the column vectors $\mathbf{S}^b$, which are used as a linear basis to form the new ensemble mean analysis state vector $\bar{\mathbf{s}}^a$. For reference, the resulting Kalman gain for the integrated RNN-ETKF is of the form,

\begin{equation}
\mathbf{K} = \mathbf{S}^b \left[ \frac{k-1}{\gamma} \mathbf{I} +  \left[ H(W_{out}(\mathbf{S}^b)) \right]^T \mathbf{R}^{-1}\left[ H(W_{out}(\mathbf{S}^b)) \right]\right]^{-1} \left[ H(W_{out}(\mathbf{S}^b)) \right]^T \mathbf{R}^{-1}.
\end{equation}

The control vector for 4D-Var can similarly be formed in the hidden/reservoir space. We use the strong constraint incremental 4D-Var, implemented using an outer and inner loop. In the outer loop, a nonlinear forecast $\mathbf{s}_t^f = \mathcal{M}_{[t,0]}(\mathbf{s}_0)$ is generated over a short optimization period, called the analysis window. In the inner loop, the linearized dynamics are used to find an improved guess for the initial state $\mathbf{s}_0$ using an iterative linear solver, and then the outer loop is repeated. If we let $\delta \mathbf{s}_0 = (\mathbf{s}_0 - \mathbf{s}_0^f)$, $\delta \mathbf{s}_0^b = (\mathbf{s}_0^b - \mathbf{s}_0^f$), and $ \mathbf{d}_t = (\mathbf{y}_t^o - H_t \circ W_{out} (\mathbf{s}_t^f))$, then the objective function is,

\begin{equation}
J(\delta \mathbf{s}_0) = J_b(\delta \mathbf{s}_0) + J_o(\delta \mathbf{s}_0),
\end{equation}

where,

\begin{equation}
J_b(\delta \mathbf{s}_0)=\frac{1}{2}(\delta \mathbf{s}_0-\delta \mathbf{s}_0^b)^T \mathbf{B}^{-1}(\delta \mathbf{s}_0-\delta \mathbf{s}_0^b),
\end{equation}

\begin{equation}
J_o(\delta \mathbf{s}_0) = \frac{1}{2} \sum_{t=0}^{N_t} \left( \mathbf{d}_t-H_t \circ  W_{out} (\mathbf{M}_{[t,0]} \delta \mathbf{s}_0) \right)^T \mathbf{R}_t^{-1} \left( \mathbf{d}_t-H_t \circ W_{out} (\mathbf{M}_{[t,0]} \delta \mathbf{s}_0) \right),
\end{equation}

\begin{equation}
\mathbf{M}_{[0,0]}  = \mathbf{I},
\end{equation}

\begin{equation}
\mathbf{M}_{[t+1,t]} = l*diag(1-\tanh(\mathbf{W}\mathbf{s}(t))^2)\mathbf{W}+(1-l)*\mathbf{I},
\end{equation}

as in equation (\ref{eqn:RNN-TLM}), and the initial condition in the original system space can be recovered by,
\begin{equation}
\mathbf{x}_0=W_{out}(\mathbf{s}_0)
.
\end{equation}

In our implementation, $H_t()$ and $W_{out}()$ are linear, so we replace them with their matrix notation. When these operators are not linear, a linear approximation via Taylor series expansion is typically applied. The minimum is found when the gradient with respect to the control vector $\delta \mathbf{s}_0$ equals 0,

\begin{equation}
\nabla_{\delta \mathbf{s}_0}{J} = \mathbf{B}^{-1}\left(\delta \mathbf{s}_0 - \delta \mathbf{s}_0^b \right) \sum_{t=0}^{N_t}\mathbf{M}^T_{[t,0]} \mathbf{H}^T_t \mathbf{R}_t^{-1}\left(\mathbf{d}_t-\mathbf{H}_t \mathbf{M}_{[t,0]}\mathbf{W}_{out} \delta \mathbf{s}_0\right)=0.
\end{equation}
We solve this by separating terms into the form `$\mathbf{A}\mathbf{x}$=$\mathbf{b}$', where $\delta\mathbf{s}_0$ is the only unknown quantity,

\begin{equation} \label{eqn:Axb}
\left( \mathbf{I} + \mathbf{B} \sum_{t=0}^{N_t} \mathbf{H}_t \mathbf{M}_{[t,0]} \right) \delta \mathbf{s}_0 = \mathbf{B} \sum_{t=0}^{N_t} \mathbf{M}_{[t,0]}^T \mathbf{H}_t^T \mathbf{R}^{-1}_t \mathbf{d}_i + \delta \mathbf{s}_0^b 
\end{equation}

and then applying the biconjugate gradient stabilized method \citep{van1992bi}. Alternative forms of equation (\ref{eqn:Axb}) are available, for example making the `$\mathbf{A}$' matrix symmetric so that the conjugate gradient method can be applied. Returning to the outer loop, a new nonlinear forecast $\mathbf{s}^f(t) = \mathcal{M}_{[t,0]}(\mathbf{s}^f_0 + \delta \mathbf{s}_0)$ is generated and the entire process is repeated with the goal of converging to the optimal nonlinear trajectory.

%We point out that the error distribution in the system space is inherited into the hidden/reservoir space due to the RNN structure. Although the RNN model is a nonlinear dynamical system, the reservoir states are transformed from system states by linear operators at each fixed time step. Hence, an additive error distribution in the original system space is equivalent to an additive error distribution in reservoir space after a linear transformation involving the matrices $\mathbf{W}_{in}$, $\mathbf{W}_{out}$, and $\mathbf{W}_{res}$. As a result of this relation, if one assumes Gaussian processes in physical space, the hidden/reservoir states are Gaussian by this inheritance. 

\subsection{Localization}
\label{sec:localization}

For the RNN model itself, scalability is enabled by partitioning the model system domain into smaller local patches, with a separate RNN trained for each patch \citep{pathak2018model}. Each local patch is assigned a small radius of `halo' points that allow information from neighboring patches as inputs to the RNN model, while computing a forecast only for the points within the patch. This follows a similar paradigm to that used for the domain decomposition of general circulation models. Localization of a geophysical forecast model is motivated by the presence of locally low dimensional chaotic dynamics \citep{oczkowski2005mechanisms}. In previous works, \citet{arcomano2020machine} demonstrated the use of RC for prediction of global scale atmospheric dynamics by applying the localization scheme described above, while \cite{lin2021} further showed that this spatial localization approach could be improved for geophysical systems by applying transformations into Fourier space. 

Localization has also been an important tool for scaling DA methods to enable application to high dimensional systems \citep{greybush2011}. The same localization procedure used to scale the RNN model can be applied in the context of DA, which provides a path to scaling the RNN-based DA methods to more realistic high-dimensional applications. We apply localization of the DA using an approach similar to the Local Ensemble Transform Kalman Filter (LETKF) \citep{hunt2007efficient}. In its original formulation, the LETKF computes a separate ETKF analysis at each model grid point, while only assimilating observations within a prescribed localization radius around that grid point. In our RNN-LETKF, we instead choose a radius around local patches. Observations are selected from an area larger than the patch itself based on the RNN localization in order to promote consistency with the analyses computed for neighboring patches. We make the design decision to maintain correspondence with the local RNN model architecture by using a radius that aligns with the input field of the local RNN, which includes the local patch and its halo points. As with the LETKF, the RNN-LETKF local analyses can be computed in parallel, after observation innovations are computed globally and distributed to each local patch. The results of the local analyses are then communicated to the neighboring patches in order to initialize the next forecast.

\subsection{Source Data for Training, Validation, and Testing}
\label{sec:source_data}

Here we describe the underlying model equations that we use to generate data for training, validating, and testing the RNN models. Lorenz (1996) developed a simple model (L96) that includes advection, dissipation, and external forcing to describe basic wavelike dynamics in the atmosphere around a latitude ring. The L96 model is a frequently used test system for DA studies \citep{abarbanel2010, penny2014hybrid, penny2017mathematical, goodliff17, chen_proactive_2019, brajard2020combining}, and multiple varieties of RNNs have been applied successfully for emulation of L96 dynamics \citep{vlachas2020backpropagation}. The L96 system is defined by a set of ordinary differential equations on a discrete finite cyclic domain,

\begin{equation} \label{eqn:L96}
f_{L96}(\mathbf{x}_i) = \frac{d\mathbf{x}_i}{dt} = \mathbf{x}_{i-1}(\mathbf{x}_{i+1}-\mathbf{x}_{i-2})-\mathbf{x}_i+F_{L96} .
\end{equation}
 
We use forcing $F_{L96}=8.0$, which is sufficient to achieve chaotic dynamics, meaning that at least one Lypaunov exponent is greater than 0. This implies initial errors will grow exponentially on average. The model is integrated with a timestep of $\delta t=0.01$ model time units (MTUs). Lorenz (1996) scaled the coefficients of the model so that the error growth over 1 MTU is roughly equivalent to 5 days in the atmosphere, relative to the state-of-the-art atmospheric models of the time. In operational prediction centers, analyses are often produced with 6-hour, 12-hour, or 24-hour update cycles, thus we will focus on DA cycles ranging up to 0.2 MTU ($\approx$ 24 hours).

\subsection{Experiment Design}
\label{sec:experiment_design}

We compare results using a variety of DA configurations. All model integrations are computed with a timestep of $\delta t$=0.01. Unless otherwise noted, we use the following parameter settings: each DA experiment is integrated for 100 MTU (or 10,000 time steps); we use observation noise of $\sigma_{noise}$=0.5 and a corresponding estimated observation error of $\sigma_{obs}$=0.5 to form $\mathbf{R}$; observations are sampled every $\tau_{obs}$=0.02 MTU ($\approx$ 2.4 hours); and we use an analysis cycle window of $\tau_{da}$=0.2 MTU ($\approx$ 24 hours). 

We use a 10-member ensemble for the RNN-ETKF, a 30-member ensemble for the RNN-LETKF applied to the higher dimensional L96-40D model configuration, and for the purposes of this discussion a 1-member `ensemble' for the RNN-4DVar. Both the RNN-ETKF and the RNN-4DVar were applied to the L96-6D model. All DA experiments are initialized by first preparing a set of perturbed spinup datasets, one for each ensemble member, applying Gaussian random noise with standard deviation $\sigma_{init}$=0.5 to the true state over a 1000 time step window (2000 for the L96-40D system). Each RNN ensemble member is synchronized with its corresponding perturbed dataset in order to produce an initial ensemble of hidden/reservoir states that reflect the uncertainty present in the noisy input data.

As noted by Lorenc (2003), additional covariance inflation is needed in the presence of model error if that error is not addressed explicitly. Covariance inflation is also typically needed due to the use of a finite ensemble size. We found an inflation parameter of 1-5$\%$ (i.e. $\gamma$ = 1.01 to 1.05) to be effective for the ETKF when applied with the `perfect' numerical model. To account for model error in the RNN, we increase the inflation parameter to 20$\%$ ($\gamma$=1.2) for the RNN-ETKF and 30$\%$ ($\gamma$=1.3) for the RNN-LETKF to account for errors in the RNN model. The 4D-Var uses an empirically chosen static error covariance that is diagonal with standard error equal to $\sigma_b = \sigma_{obs}$, an analysis time at the start of the window, and uses 2 outer loops.

To mimic a realistic scenario of geophysical prediction, we focus mainly on cases where the variables are sparsely observed. If not otherwise stated, the L96-6D model is observed only at the first, second, and fourth nodes. For the L96-40D model, we limit the observing network to only 15 nodes.

\section{Results} 

\subsection{Assessment of error growth rates}

\begin{figure}[!htpb]
    \centering
    \includegraphics[width=0.9\textwidth]{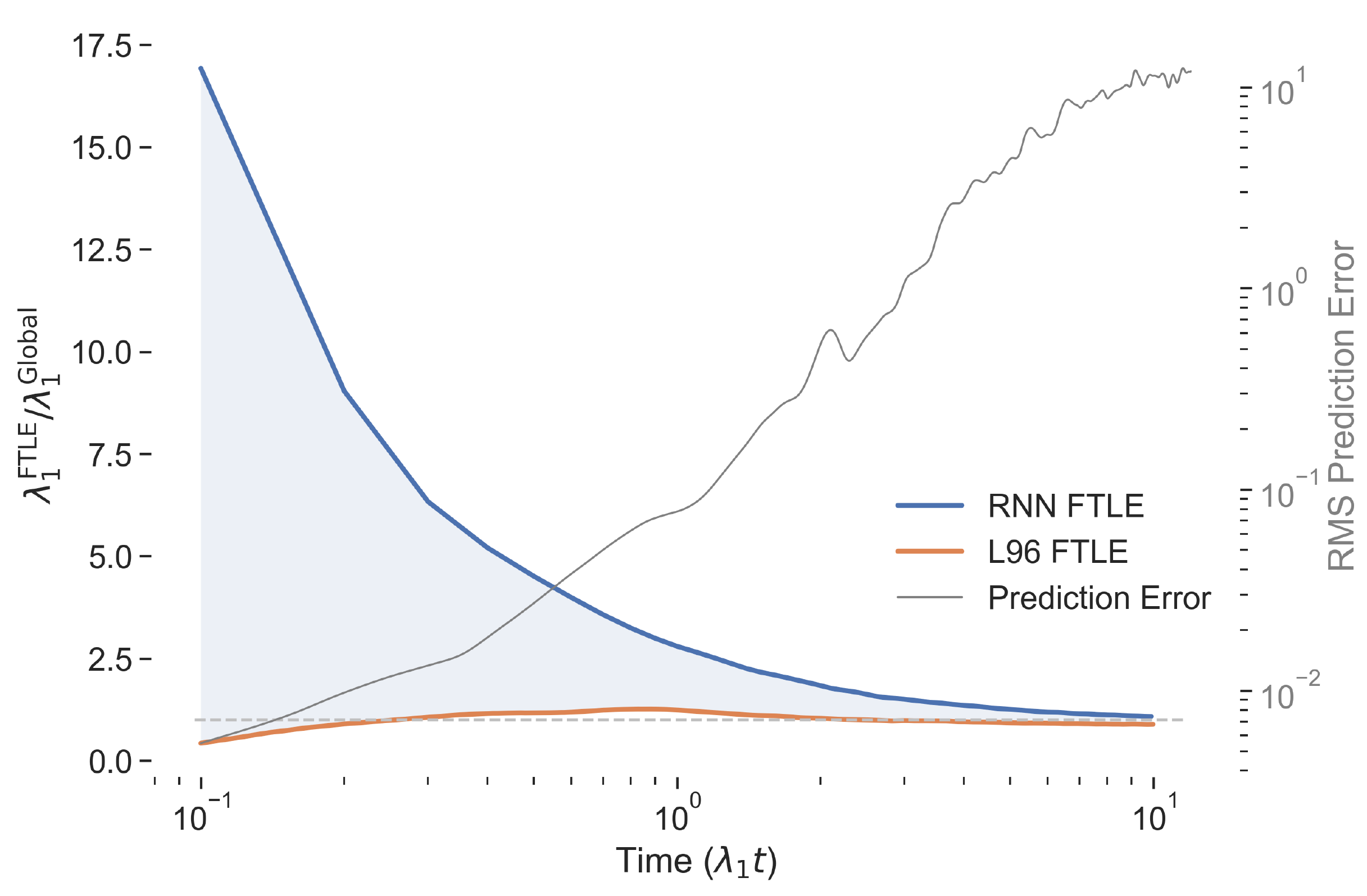}
    \caption{Convergence of the leading FTLE ($\lambda_1$) for a trained RNN (model 1 in Table \ref{tab:macro-params}) averaged over 100 initial conditions of the L96-6D system. As the RNN is integrated for longer periods of time, the error growth rates generated by the RNN model become more accurate. However, over the same period there is an exponential growth of errors in initial conditions.}
    \label{fig:FTLE}
\end{figure}

\begin{figure}[!htb]
    \centering
    \includegraphics[width=1.0\textwidth]{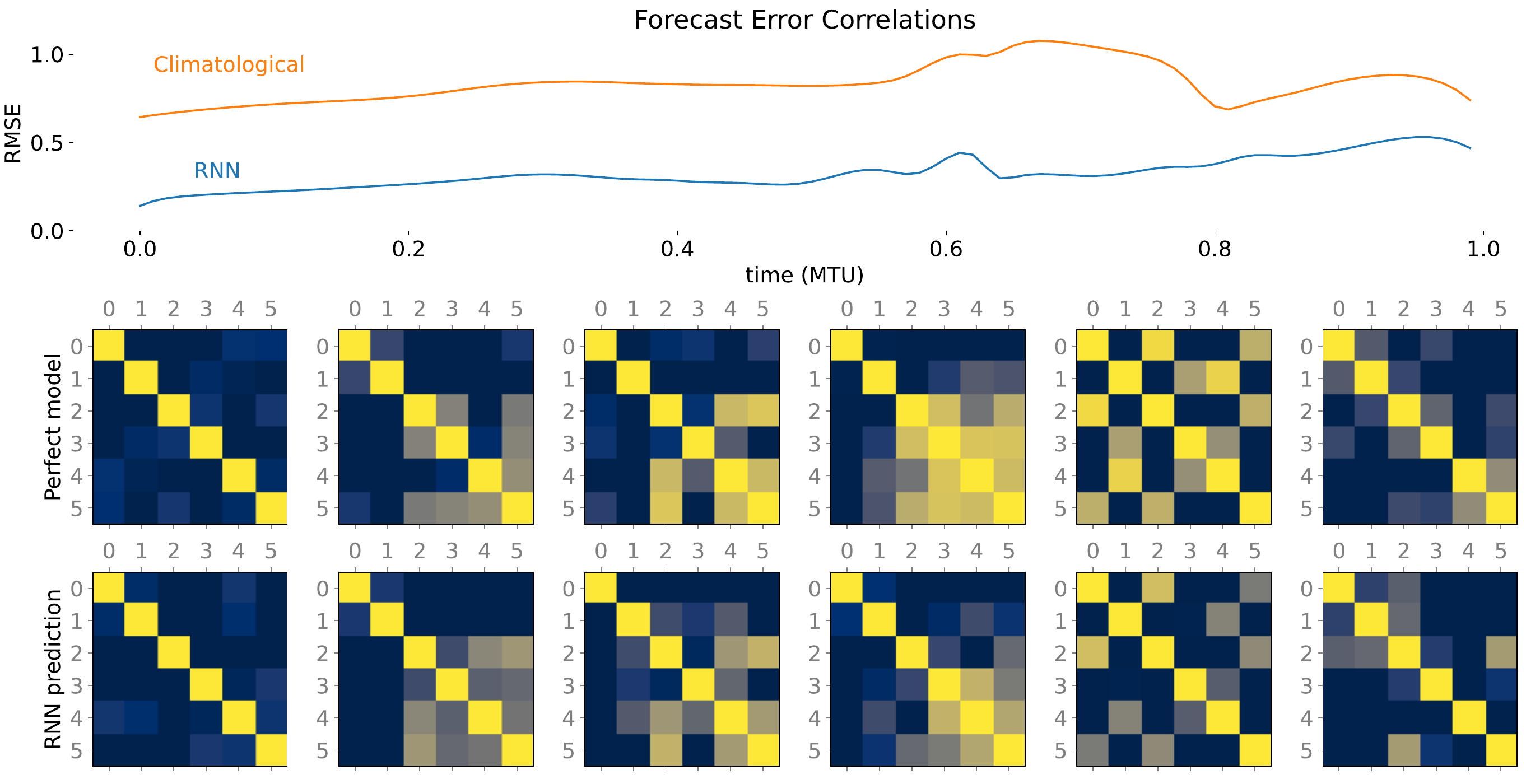}
    \caption{(Top) RMSE of the forecast error correlation matrix over time for the L96-6D system, comparing an example RNN ensemble forecast (blue) to the climatological error correlation matrix (orange), both evaluated versus a 100-member ensemble forecast using the perfect reference model. The initial conditions are sampled from the test dataset, and initial ensemble perturbations are generated using a Gaussian distribution with standard deviation 0.1. (Bottom) Forecast error correlation matrix for the reference perfect numerical model and the RNN model over the same forecast period, valid at times 0.0, 0.2, 0.4, 0.6, 0.8, and 1.0. The color scale ranges from 0 to 1.}
    \label{fig:RNNfcsterrcorr}
\end{figure}

An essential consideration of DA is the behavior of long and short term error growth, which can be characterized by the Lyapunov exponent (LE) spectrum and finite-time (also known as `local') Lyapunov exponents (FTLEs)~\citep{abar92,abar96}. If one considers DA as the synchronization of a model with the natural process from which measurements are drawn, then the conditional LE spectrum of this coupled model-nature system must be driven negative to ensure the model synchronizes with the observed system \citep{penny2017mathematical}. Previous studies \citep{pathak2018model, griffith2019forecasting, platt2021forecasting} have already shown that reservoir computing can be used to reproduce the Lyapunov spectrum of the source system. This spectrum characterizes the long time average exponential growth rates of small errors in the system trajectory. However, at very short timescales, we find that the error growth rates of our trained RNNs are not well representative of the error growth rates produced by the source system dynamics. We find instead that the FTLEs of the RNN dynamics converge towards the Lyapunov exponents of the source system dynamics over a transient period of a few Lyapunov timescales (Figure \ref{fig:FTLE}). While the growth rates of errors in the RNN models take some time to converge to the true growth rates, the forecast error correlations appear to be estimated relatively accurately at short lead times (an example is shown in Figure \ref{fig:RNNfcsterrcorr}). This indicates that while it may be desirable to improve the convergence rates of the FTLEs produced by the RNN, the effect can be compensated for by using a scalar multiplicative inflation applied to the forecast error covariance matrix.

\subsection{Control RNN case}

To demonstrate the need for DA, we commence our cycled forecast experiments with a control case that sets a baseline for the performance of the RNN without using DA. Here, observations are inserted directly into the RNN  as defined by as in equations (\ref{eqn:RC}) and (\ref{eqn:RC2}). If the system is fully observed, then this amounts to replacing $\mathbf{x}(t_i)$ in equation (\ref{eqn:RC}) with observed data.
% via, 
%\begin{equation}
%\mathbf{s}^a = \mathbf{s}^b + %H^T(\mathbf{y}^o - H\mathbf{s}^b) ,
%\end{equation}
%which does not incorporate error %covariances.

\begin{figure}[!htb]
    \centering
    \includegraphics[width=1.0\textwidth]{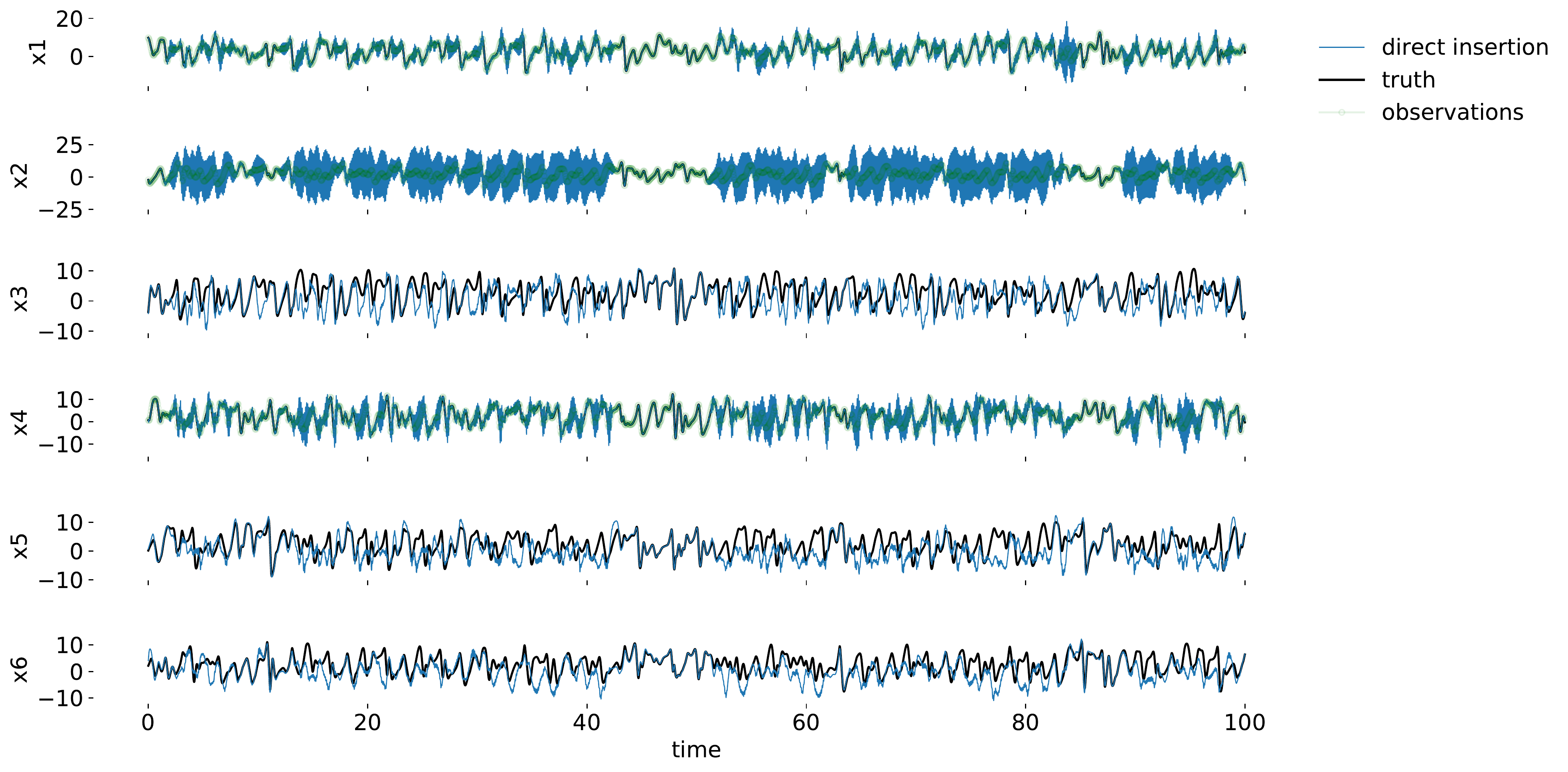}
    \caption{Direct insertion using the RNN with perfect ($\sigma_{noise}$=0) observations of the system at points (1,2,4), sampled with $\tau_{obs}$=0.05 ($\approx$6 hours), which is every 5 model time steps. The RNN alone cannot successfully recover the true trajectory when observations are sparse and noisy. This figure provides a corresponding entry in figure \ref{fig:RMSEcontrol}.}
    \label{fig:RNNcontrolexample}
\end{figure}

\begin{figure}[!htb]
    \centering
    \includegraphics[width=1.0\textwidth]{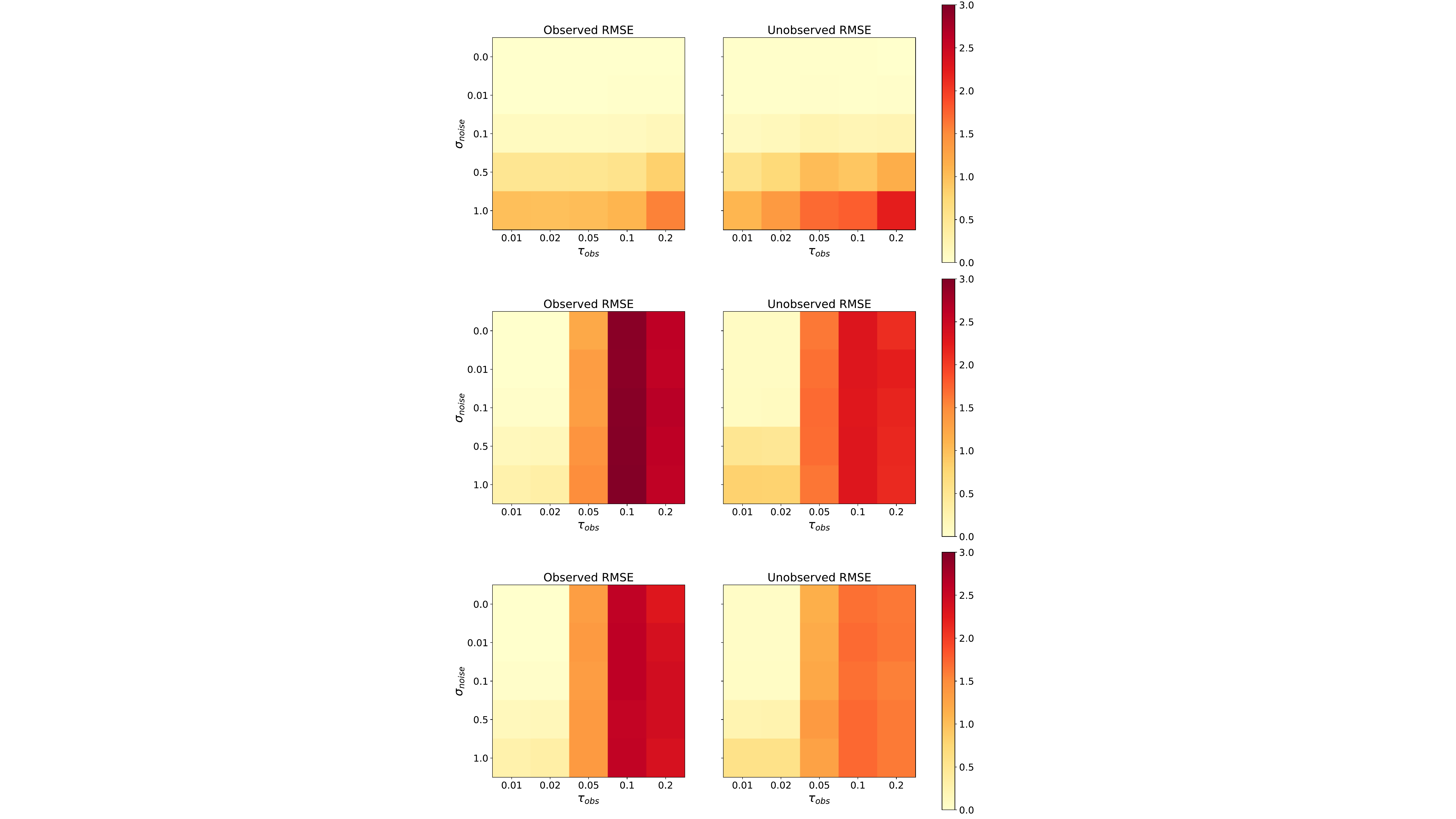}
    \caption{Normalized RMSE using direct insertion using (Top) the `perfect' numerical model and (Bottom) the RNN model 1. Both are applied to the L96-6D system integrated over 100 MTU, varying the observation noise $\sigma_{noise}$ and the observation timestep $\tau_{obs}$. (Left) RMSE of observed points (indexes 1,2,4). (Right) RMSE of points that are not observed (indexes 3,5,6). A normalized RMSE of 1.0 equals the L96 system's climatological standard deviation. Note that the conventional model is more sensitive to increased noise, while the RNN model is more sensitive to the observing frequency.}
    \label{fig:RMSEcontrol}
\end{figure}

\citet{lu2017reservoir} examined a ‘sparse in space’ case that limited the forcing of an RC model to only a subset of inputs. Their results showed that synchronization of the full state can be achieved even when observing only a subset of the variables of the system. However, we find that this type of direct insertion method for synchronization fails as the observations become more sparse in time (see Figure \ref{fig:RNNcontrolexample} and the middle and bottom rows of Figure \ref{fig:RMSEcontrol}). 

Before further evaluating the RNN, we first consider for reference the case of direct insertion of observations into the original L96-6D numerical model, using the update equation,

\begin{equation}
\mathbf{x}^a = \mathbf{x}^b + \mathbf{H}^T(\mathbf{y}^o - \mathbf{H}\mathbf{x}^b).
\end{equation}
As should be expected, providing perfect observations of all variables at every model time step ($\delta t$=0.01) produces exact synchronization between the driver signal and the numerical model trajectory. Increasing the timestep between observations as high as 0.2 MTU does not significantly degrade the state estimates, with errors peaking at ~2.5e-7. Increasing the observational noise generally increases the error in the state estimates by a similar magnitude. When the number of observed variables is reduced (e.g. to 3 or 2 out of 6), the system experiences transient synchronization with occasional bursting. While still using perfect observations, combining reduced observations (e.g. 50\%) and using longer timesteps (e.g. $\delta t$=0.1) actually improves stability compared to using a time step of $\delta t$=0.01, and leads to synchronization. However, when observation noise is added to this combination of sparseness in space and time, the bursting phenomena return, particularly in the unobserved variables (see top row of Figure \ref{fig:RMSEcontrol}).

The situation is quite different with the RNN model. Increasing the time step between the (noise-free) observations significantly degrades the RNN estimates, first adding high frequency oscillations (e.g. with $\delta t$=0.02 to $\delta t$=0.1), and then leading to trajectories with little discernible connection to the L96 dynamics (e.g. at $\delta t$=0.2). Recall that for these experiments the RNN is trained on data that have a temporal resolution matching the underlying model timestep $\delta t$=0.01. Reducing the frequency of the input driving signal allows the hidden/reservoir state to drift. Observing frequently ($\delta t$=0.01) but removing the observation of one variable degrades the estimates of that variable without noticeably affecting the rest, while removing the observation of more than one variable can cause occasional degradation of the remaining observed variables. Due to the presence of the hidden/reservoir state, which maintains a memory of the past trajectory, the RNN itself is relatively insensitive to the introduction of noise to the observations. When observing the full system state, noisy observations supplied as driving data to the RNN simply increase high frequency noise in the analyzed state estimate, without resulting in divergence between the RNN and the true signal. Combining any of these constraints on the RNN appears to have additive effects. A comparison of the errors in the RNN using direct insertion with a range of observation noise and observing frequency (of which Figure \ref{fig:RNNcontrolexample} is one instance) is shown in Figure \ref{fig:RMSEcontrol}. 

We note that even for this simple L96 model, the total set of direct insertion experiments using RNN model 1 was about 10\% faster, and RNN model 2 about 30-40\% faster, than the total set of direct insertion experiments using the conventional numerical integration of the L96 differential equation (\ref{eqn:L96}). We do not claim that these results can be easily extrapolated to other applications. However, we do emphasize that the projection of the system dynamics to the higher dimensional hidden/reservoir state does not necessarily imply that the computations become more costly. 

To summarize - simply providing the sparse and noisy observation data directly to the RNN is not adequate for initializing forecasts, which provides motivation for the use of a more sophisticated DA strategy.

\subsection{RNN-DA with sparsely observed dynamics}

DA methods provide most of their value when observations are sparse and noisy. Here, we restrict the observing network to only the first, second, and fourth nodes of the 6-dimensional cyclic L96 system (L96-6D). This leaves two patches that are unobserved for the duration of each experiment - the third node and the combined fifth and sixth node. We noticed no qualitative differences in other configurations of the observing system layout at the same 50$\%$ coverage level.

\begin{figure}[!htb]
    \centering
    \includegraphics[width=1.0\textwidth]{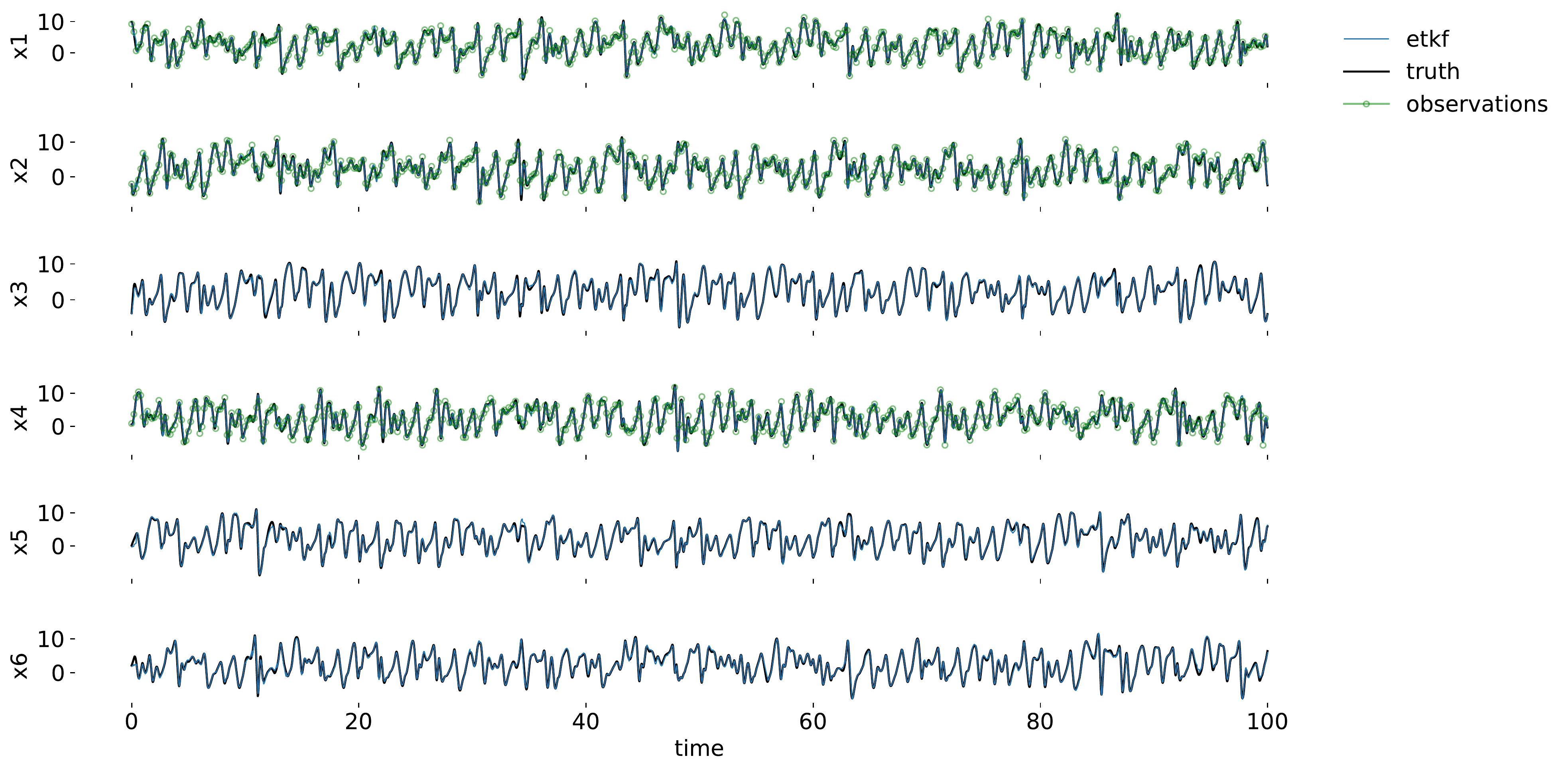}
    \caption{The RNN-ETKF, assimilating observations at only three points (1,2,4) at increments of $\tau_{obs}$=0.2 (i.e. every 20 model time steps), converges to the true system trajectory within a few timesteps. Note the true and estimated trajectories are nearly indistinguishable. This figure provides a corresponding entry in figure \ref{fig:etkfsweep}.}
    \label{fig:ETKFanalysis}
\end{figure}

\begin{figure}[!htb]
    \centering
    \includegraphics[width=1.0\textwidth]{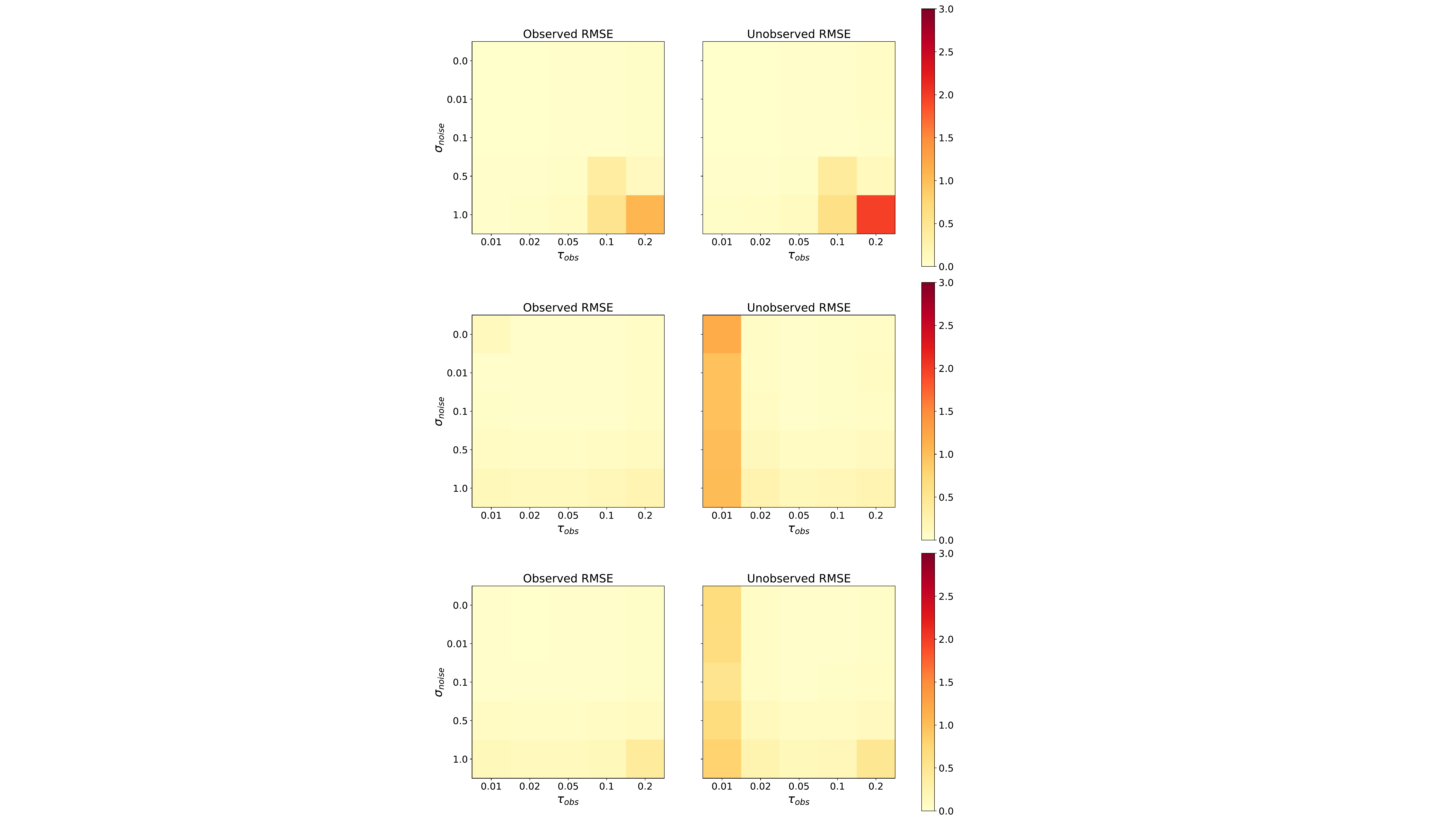}
    \caption{Normalized RMSE of (Top) conventional ETKF using the `perfect' numerical model, (Middle) the RNN-ETKF using RNN model 2 (hidden/reservoir dimension 800), and (Bottom) the RNN-ETKF using RNN model 1 (hidden/reservoir dimension 1600). All are applied to the L96-6D system integrated over 100 MTU, varying the observation noise $\sigma_{noise}$ and the observation time step $\tau_{obs}$. The analysis cycle is adjusted for each case so that $\tau_{DA}$=$\tau_{obs}$. (Left) RMSE of observed points (indexes 1,2,4). (Right) RMSE of points that are not observed (indexes 3,5,6). Surprisingly, the RNN-ETKF outperforms the conventional ETKF (which uses the perfect model) when both the observational noise and observing timestep are large.}
    \label{fig:etkfsweep}
\end{figure}

We first implement the RNN-ETKF and compare to the conventional ETKF using the `perfect' numerical model. We find that with the exception of the case in which analyses are updated frequently ($\tau_{DA}$=$\tau_{obs}$=$\delta t=0.01$), the RNN-ETKF using both RNN models 1 and 2 performs quite well in comparison (Figures \ref{fig:ETKFanalysis} and \ref{fig:etkfsweep}). We note that the total set of ETKF experiments using RNN model 1 had about equal run time, while RNN model 2 was about 25\% faster, compared to the run time of the total set of ETKF experiments using the conventional numerical integration of the L96 differential equation (\ref{eqn:L96}).

\begin{figure}[!htb]
    \centering
    \includegraphics[width=1.0\textwidth]{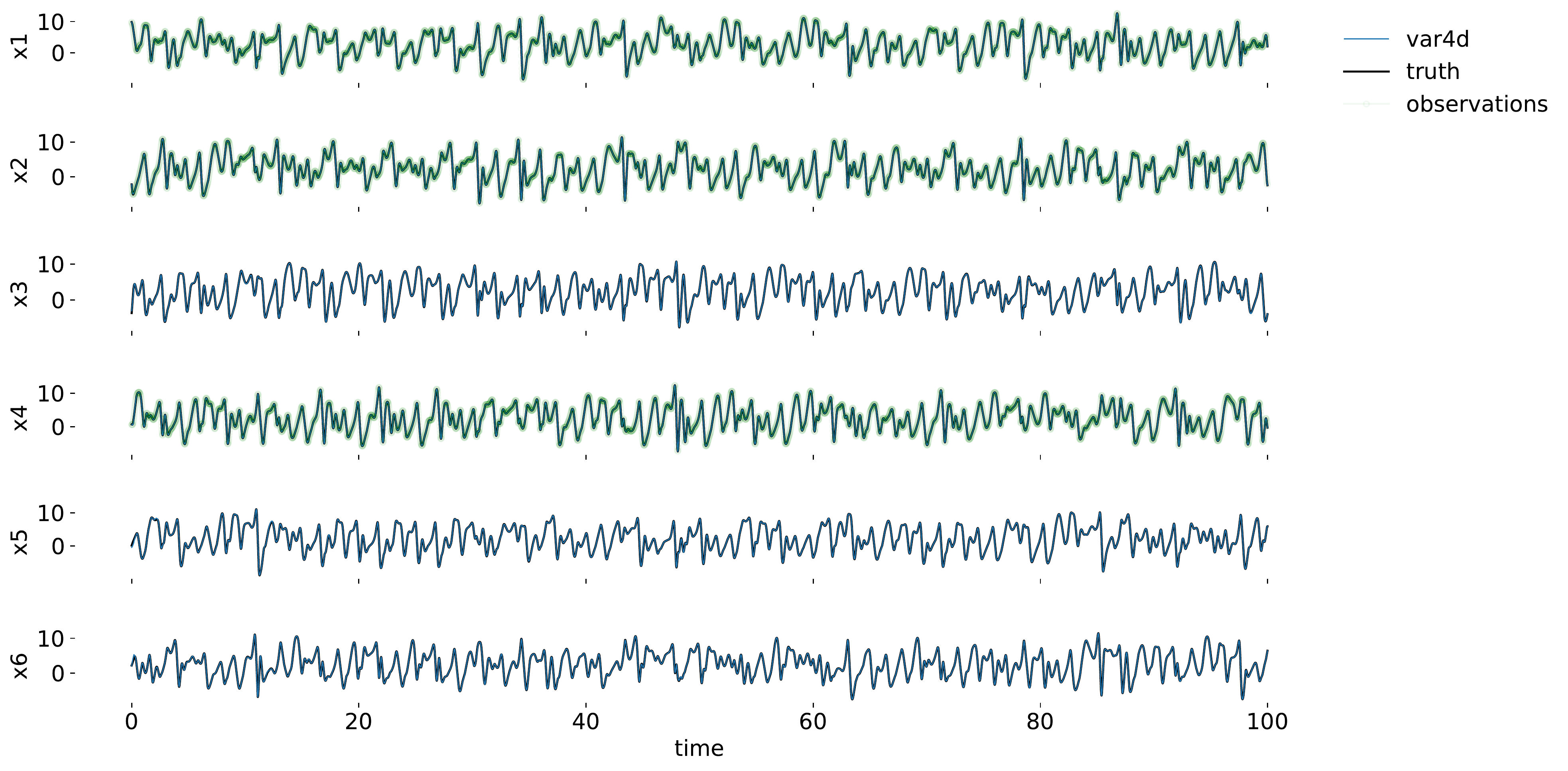}
    \caption{The RNN-4DVar, assimilating observations at only three points (1,2,4) at increments of $\tau_{obs}$=0.02 (i.e. every 2 model time steps), with observation noise set to $\sigma_{noise}$=0. The analysis cycle is $\tau_{obs}$=0.2 (i.e. every 20 model time steps). This figure provides a corresponding entry in figure \ref{fig:var4dsweep}.}
    \label{fig:4DVARanalysis}
\end{figure}

\begin{figure}[!htb]
    \centering
    \includegraphics[width=1.0\textwidth]{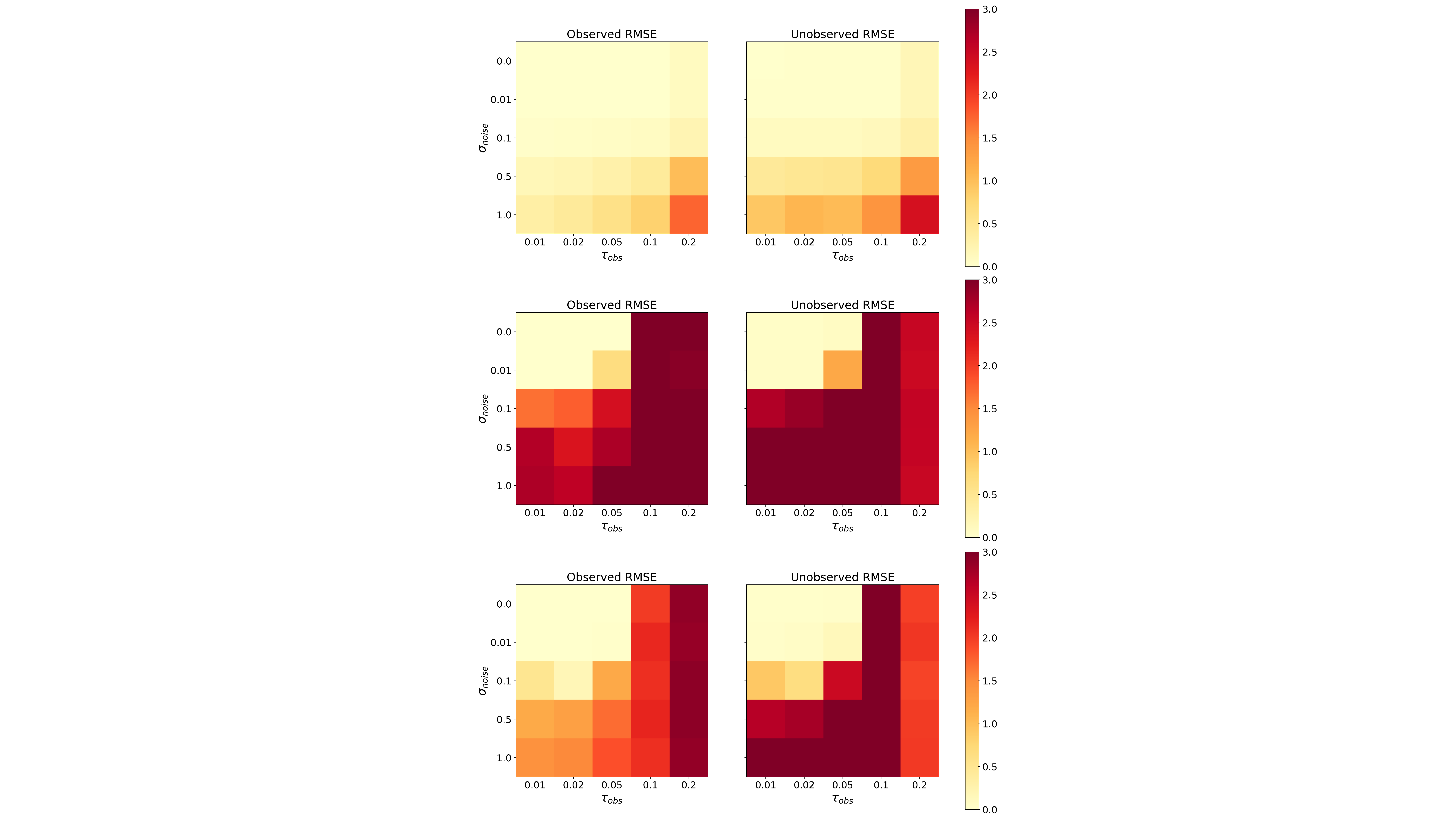}
    \caption{Normalized RMSE of (Top) conventional 4D-Var using the `perfect' numerical model, (Middle) the RNN-4DVar using RNN model 2 (hidden/reservoir dimension 800), and (Bottom) the RNN-4DVar using RNN model 1 (hidden/reservoir dimension 1600). All are applied using an analysis cycle of $\tau_{DA}$=0.2 (i.e. every 20 model time steps). The RNN-4DVar performs best with frequent observations and relatively low observational noise, but otherwise has degraded performance.}
    \label{fig:var4dsweep}
\end{figure}

The RNN-4DVar method performs well when the observational noise is small, but is sensitive to increasingly sparse and noisy observing sets (Figures \ref{fig:4DVARanalysis} and \ref{fig:var4dsweep}). As the underlying RNN model is improved (from model 2 to model 1 in Table \ref{tab:macro-params}), this appears to improve the performance of the 4D-Var correspondingly. The sensitivity of the RNN-4DVar to observational noise may be exacerbated by errors in the RNN model equations from which the TLM and adjoint operators are derived, and also the approximated background error covariance matrix. A further drawback is that the experiments using the RNN-4DVar required 1-2 orders of magnitude more computational time than the conventional 4D-Var applied using the numerical integration of the L96 differential equation (\ref{eqn:L96}) and its TLM and adjoint equations.

The difference between FTLEs estimated by the RNN and the numerical model at short lead times indicates that the linearized dynamics (i.e. the TLM and adjoint) are not well represented at these timescales. The RNN models used here generally under-represent the magnitude of error growth at short timescales. This affects the ETKF as well but is alleviated by the application of multiplicative inflation, and gives some explanation for why the ETKF is more stable than 4D-Var. The strong-constraint 4D-Var used here assumes a `perfect' model. We expect that transitioning from the strong-constraint 4D-Var formulation to the weak-constraint 4D-Var approach should further improve the RNN-4DVar performance, as the latter explicitly accounts for errors in the model.

\subsection{Scaling RNN-DA to higher dimensions}
\label{sec:scaling_up}

Next we demonstrate that the components of the RNN-DA system can scale as the system size increases. Given the results of the previous section, we will focus the RNN-ETKF. We increase the dimension of the L96 system from 6 to 40 in order to demonstrate the scalability of the system via the aforementioned localization scheme. The system is partitioned into 20 local subgroups of 2 points each, and a separate RNN is trained for each local subgroup. The input signal to each RNN is made up of the 2 points in its associated local subgroup, plus 4 neighboring points on either side, giving each RNN an input dimension of 10 and an output dimension of 2. For each local subgroup, observations are assimilated if they are located within the range of the local RNN input domain. In this example we observe 15 nodes of the system, leaving 25 nodes unobserved.

\begin{figure}[!htb]
    \centering
    \includegraphics[width=\textwidth]{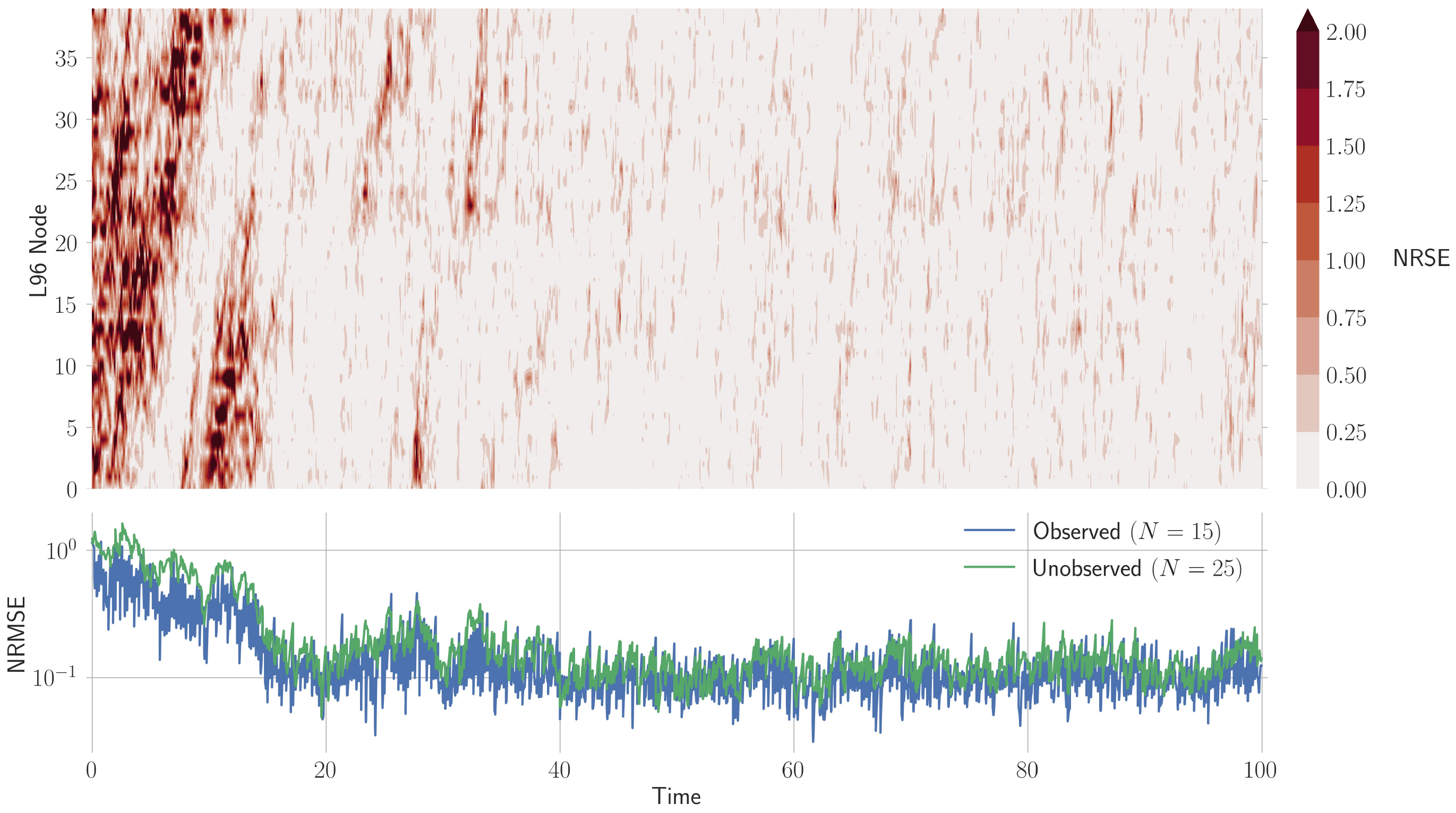}
    \caption{Normalized error of the RNN-LETKF based state estimation for the L96-40D system using RNN Model 3. (Top) Normalized Root Square Error (NRSE) shown for each node of the L96 system (y-axis). (Bottom) NRMSE computed separately for the observed and unobserved nodes of the Lorenz system, with 15 nodes of the system observed. Note the y-axis is logarithmic in the lower plot. The error in both plots are normalized by the temporal standard deviation of the true trajectory. The RNN-LETKF uses a 30 member ensemble, with $\sigma_{obs}$=$\sigma_{noise}$=0.5,  and macro-scale parameters indicated by RNN model 3 in Table \ref{tab:macro-params}. The observed nodes of the system are $[0,3,5,8,10,14,16,19,20,25,27,30,34,36,39]$.}
    \label{fig:rc_letkf_error}
\end{figure}

\begin{figure}[!htb]
    \centering
    \includegraphics[width=\textwidth]{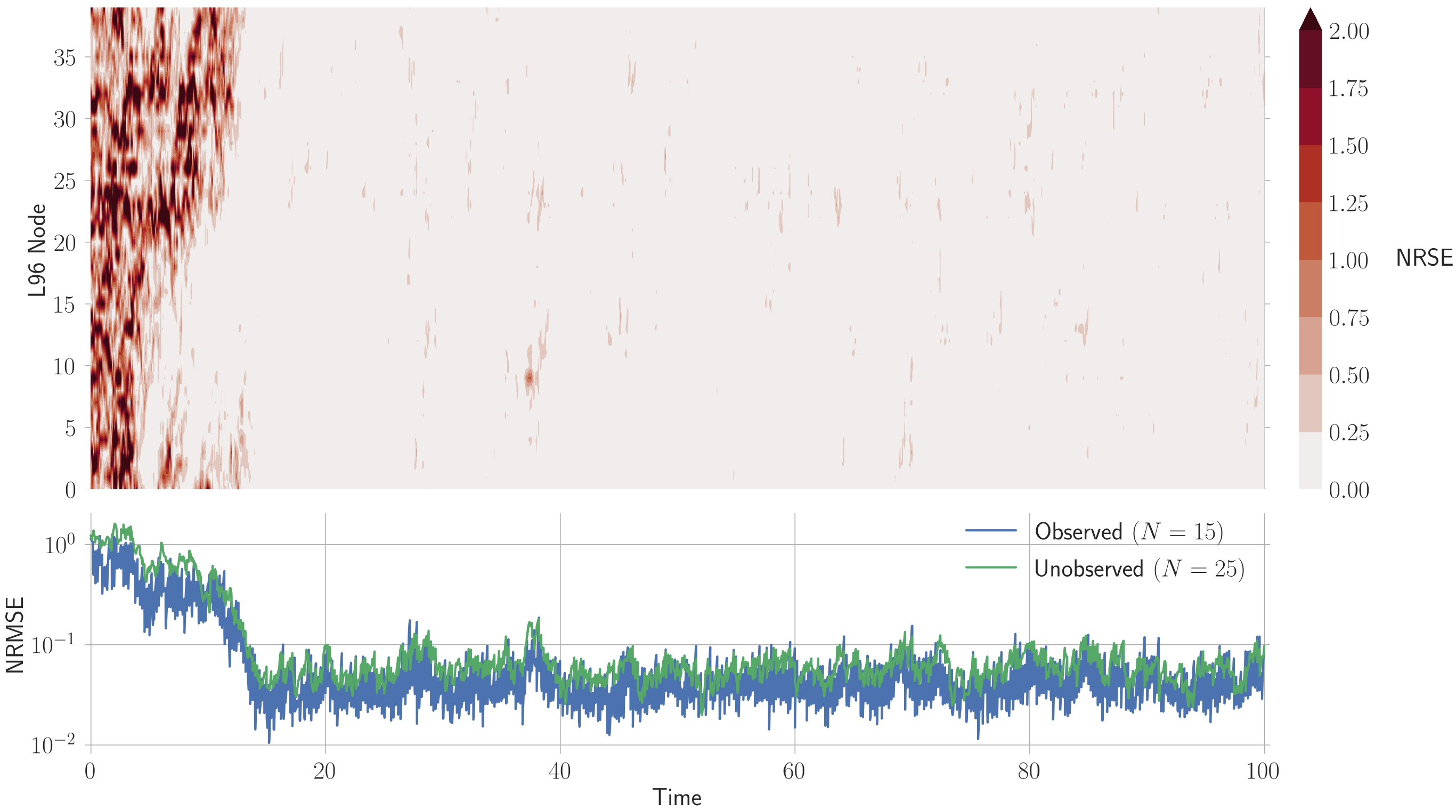}
    \caption{As in Figure \ref{fig:rc_letkf_error}, but with $\sigma_{noise}$=0.1.}
    \label{fig:rc_letkf_low_noise}
\end{figure}

The Normalized Root Square Error (NRSE) of the trajectory estimated by the RNN-LETKF is shown in Figure \ref{fig:rc_letkf_error} using observation noise $\sigma_{noise}$=0.5, and in Figure \ref{fig:rc_letkf_low_noise} with reduced observation noise $\sigma_{noise}$=0.1. An additional perturbation is added at the first timestep to ensure that the background estimate of the system state is far from truth. After roughly 20 MTU, the cycled DA system converges and provides an accurate estimation of the system trajectory. The lower panels in Figures \ref{fig:rc_letkf_error} and \ref{fig:rc_letkf_low_noise} show the Normalized Root Mean Square Error (NRMSE) of the observed and unobserved nodes of the L96-40D system separately. The RNN-LETKF system appears to provide an accurate transfer of information from the observed to the unobserved variables. We note that producing accurate predictions for the L96-40D system requires the use of a larger hidden/reservoir dimension, a larger training dataset, and a longer spinup time. The results shown here use a 6000-dimensional reservoir for each local subgroup and 200,000 time steps of training data, with each local reservoir trained independently and in parallel. We note that with much longer training datasets, the training of the micro-scale parameters can easily be applied in batches with equivalent results, with the macro-scale parameters estimated from a sample of batches.

\section{Conclusions}

One of the most common procedures in the field of data assimilation (DA) is to combine a computational model with observations to estimate the state of a partially observed system. This procedure is used for applications such as initializing models to make real-time forecasts, or creating historical reconstructions based on a limited archive of observation data. An essential element of modern DA algorithms is the expectation that the forecast model responds accurately to small perturbations in the initial conditions. By integrating recurrent neural networks (RNN) with the ensemble Kalman filter and 4D-Variational DA methods, we have demonstrated that RNNs can produce reasonable representations of the system response to uncertainty in initial conditions. Critical to this demonstration was the assimilation of sparse observation data, which requires sufficiently accurate ensemble forecast error covariance statistics and tangent linear model dynamics to propagate information from observed to unobserved variables.

Comparing the application of a DA method using RNN-based forecast models to the same DA method using a ‘perfect’ model provides a useful analogue to the real-world scenario in which an imperfect numerical model is used to estimate the true state of a natural system. Beyond the practical applications that a ML model may have, there is also much to be learned from determining the necessary elements for ML models to be used applications like reanalysis and operational numerical weather prediction. One example provided was the reproduction of the finite-time Lyapunov exponents (FTLEs).

We note that while the RNN-DA methods have been applied to training data generated from a known model, as long as adequate training data exists the methods apply equally to systems for which no known theoretical or computational model is available. Further, the RNN-4DVar can be produced for models in which the tangent linear and adjoint models are either not available or are too difficult to calculate. The RNN-4DVar could easily be implemented in hybrid forms in which a conventional numerical model is used for the outer loop and RNN-based tangent linear model is used for the inner loop, or alternatively the RNN could be used as the nonlinear model in the outer loop to reduce computational costs while still using a numerical tangent linear model and adjoint in the inner loop. We also note that while the RNN models were trained once on historical data and held fixed during the RNN-DA cycling, it is straightforward to perform online retraining of the RNN model during the DA cycle, e.g. as suggested by \cite{brajard2020combining}.

Further methods are needed to optimize the design and training of the ML model used in this study to replace the numerical forecast model. More development is needed in ML modeling of chaotic dynamics to ensure a rapid convergence of the finite time Lyapunov exponents toward the true values. While a simplified RNN resembling the reservoir computing approach was applied here, we expect that our approach could be applied with more sophisticated types of RNN that also use hidden state representations, such as Long Short Term Memory (LSTM) or Gated Recurrent Unit (GRU) architectures.

%\newpage
\appendix
\section{Details of the RNN Training}

Bayesian optimization is an optimization technique that ``minimizes the expected deviation from the extremum'' of a target loss function \cite{Mockus75}. The technique is useful mainly for global optimization of expensive nonlinear functions for which no gradient is computed. The specific algorithm used here is the efficient global optimization algorithm described by \cite{jones1998efficient} and implemented by \cite{SMT2019}, using a Kriging surrogate model. In short, the algorithm starts by sampling a number of initial points over the search space. It then fits a Gaussian process regression across those points, enabling interpolation and extrapolation. After the fit, the algorithm computes the ``expected improvement'' of searching a new region of the space and then chooses points based on maximizing this criterion.

The loss function chosen is based on computing the scaled mean squared error (MSE) of long-range predictions over the test data set, as described by equation (\ref{eq:macro_loss}). Hyperparameters of this optimization are shown in Table \ref{tab:hyperparams}, and include the length of the training data for the RC, the number of forecasts in the validation set, and the length of those forecasts over which the MSE is computed. The following options for the Bayesian optimization algorithm are kept fixed: the number of iterations of the algorithm (15), the number of parallel samples computed (4), and the number of optimization start points (100).

\begin{table}
\caption{\label{tab:hyperparams}Hyperparameters used for the Bayesian optimization. For model 3, we perform the optimization using a reduced reservoir size (2,000) due to the computational cost of the algorithm. Predictions are made with a larger reservoir (6,000), see text for details.}
\begin{center}
 \begin{tabular}{c c c c} 
 \hline
 Hyperparameter & RNN model 1 & RNN model 2 & RNN model 3 \\ [0.5ex] 
 \hline\hline
 hidden/reservoir dimension & 1600 & 800 & 2,000 (6,000) \\
 \hline
 size of training set & 100,000 & 100,000 & 200,000 \\
 \hline
 number of long forecasts (M) & 100 & 100 & 100 \\
 \hline
 length of long forecasts (N) in MTU & 10.0 & 10.0 & 10.0 \\
  \hline
 sparsity of reservoir matrix  & 99\% & 99\% & 99\% \\
  \hline
 input weighting ($\sigma$) limits  & [0.001, 1.0] & [0.001, 1.0] & [0.001,1.0] \\
  \hline
 leak rate ($l$) limits & [0.001, 1.0] & [0.001, 1.0] & [0.001, 1.0] \\
  \hline
 spectral radius ($\rho$) limits & [0.1, 1.5] & [0.1, 1.5] & [0.1, 1.5] \\
  \hline
 Tikhonov parameter ($\log \beta$) limits & [log(1e-8), log(1.0)] & [log(1e-8), log(1.0)] & [log(1e-8), log(1.0)] \\
% [1ex] 
 \hline
\end{tabular}
\end{center}
\end{table}

All RNNs here use a sparsity of 99\% (density of 1\%) for the reservoir adjacency matrix $\mathbf{W}_{res}$. For L96-6D experiments, we use the following hyperparameters for the Bayesian optimization: $M$=100 initial points for validation forecasts, chosen randomly without replacement; a forecast length of 10.0 MTU, or $N$=1000, equal to $\approx$10 Lyapunov timescales. The macro-scale parameters learned from the Baysian optimization process are provided in Table (\ref{tab:macro-params}).

The Bayesian optimization algorithm is computationally intensive for Model 3, as each iteration of the algorithm requires training the micro-scale parameters for 20 localized RC models and evaluating their forecast skill. \cite{lukosevicius_practical_2012} suggested that using a reduced reservoir size while searching for optimal hyperparameters is an effective means to reduce computational costs. \cite{vlachas2020backpropagation} also showed that increasing the reservoir size for a localized RC model trained on the L96 system with fixed macro-scale parameters simply increases the valid prediction time. Thus, to make the training of Model 3 more tractable, we use a reduced reservoir size within the Bayesian optimization algorithm to identify the best-performing macro-scale parameters (Table \ref{tab:macro-params}). We use a hidden/reservoir dimension of 2,000 during the optimization, while for all Model 3 forecasts used in DA experiments and statistical tests we use the larger hidden/reservoir dimension of 6,000 (Figures \ref{fig:rc_letkf_error}, \ref{fig:rc_letkf_low_noise}, and \ref{fig:hist3}). 

\begin{table}
\caption{\label{tab:macro-params}Trained values of the macro-scale parameters for the RNNs used as forecast models by the data assimilation algorithms.}
\begin{center}
 \begin{tabular}{c c c c} 
 \hline
 Trained Parameter & RNN model 1 & RNN model 2 & RNN model 3 \\ [0.5ex] 
 \hline\hline
 spectral radius ($\rho$) & 0.10036271 & 0.10000000 & 0.34378377 \\
 \hline
 input weighting ($\sigma$) & 0.06627321 &   0.05343709 & 0.05219330 \\
 \hline
 leak parameter ($l$) & 0.70270733 & 0.69460913 & 0.40813549\\
 \hline
  Tikhonov parameter ($\beta$) & $\exp$(-18.41726026) & $\exp$(-14.33030495) & $\exp$(-12.53138825) \\
% [1ex] 
 \hline
\end{tabular}
\end{center}
\end{table}

\begin{figure}
    \centering
    \includegraphics[width=.8\textwidth]{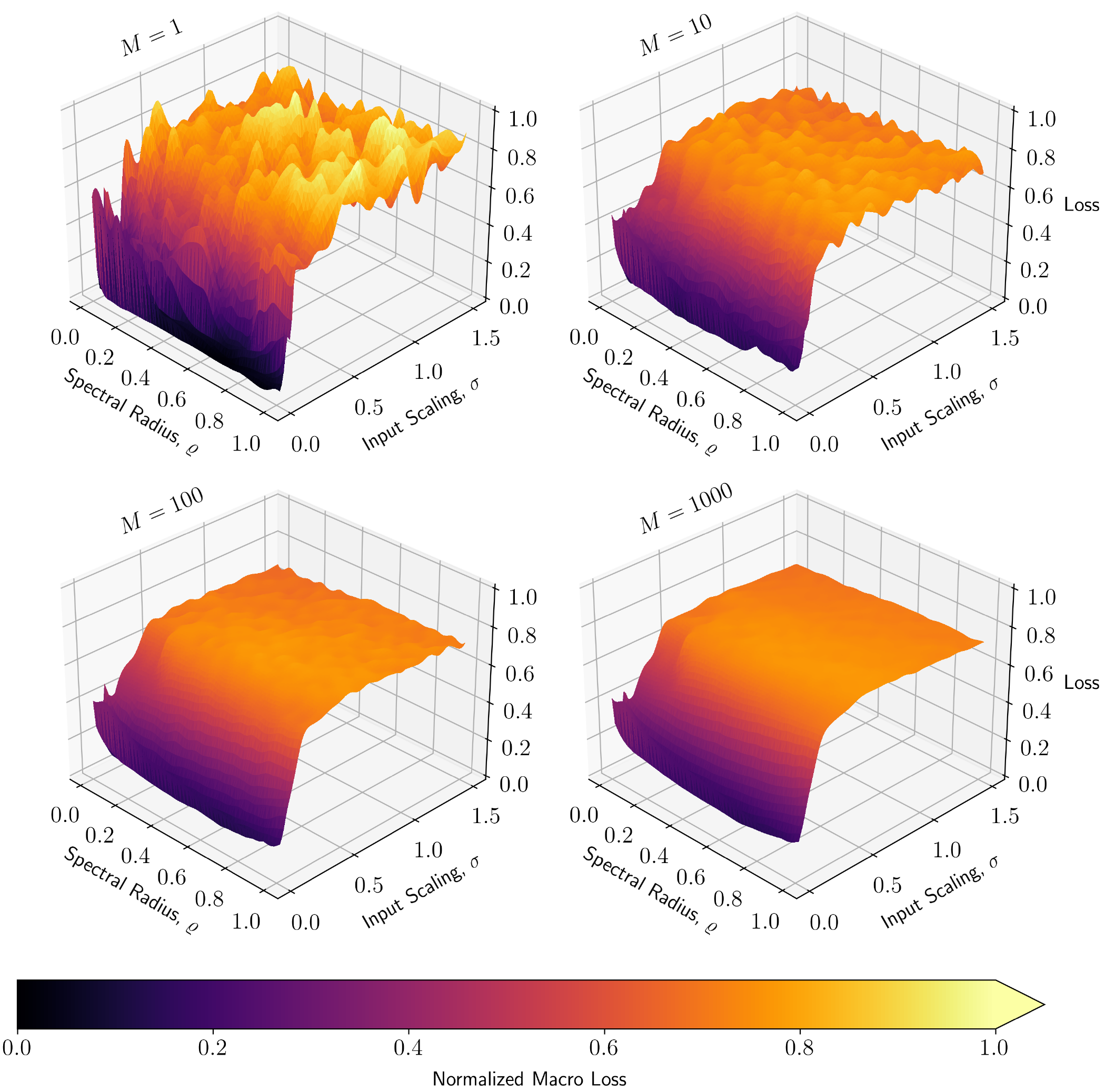}
    \caption{Normalized macro loss function, $\mathscr{L}_{macro}$ (see equation \ref{eq:macro_loss}) for various $M$ visualized in the $(\sigma,\rho)$ plane. Increasing $M$ regularizes the loss function, revealing an approximate global minimum in parameter space. The loss function shown here is based on an 800 dimensional hidden/reservoir state RNN model prediction of the L96-6D system.}
    \label{fig:cost_landscape}
\end{figure}

A visualization of the macro loss function landscape in the $(\sigma,\rho)$ plane is shown in Fig. \ref{fig:cost_landscape} based on RNN forecasts of the 6 dimensional Lorenz 96 model (described in Section \ref{sec:source_data}). When $M$=1, the macro loss function exhibits many local minima. This indicates that the resulting trained RNN may produce accurate forecasts for a particular set of initial conditions, but does not generalize well to other points on the system attractor. Evaluating forecasts from a greater number of initial conditions results in a smoother loss function landscape. However, increasing $M$ also carries a corresponding increase in cost for the evaluation of the macro loss function. We select $M$=100 to balance the computational cost and the diminishing returns observed beyond this point.

\begin{figure}
    \centering
    \includegraphics[width=0.8\textwidth]{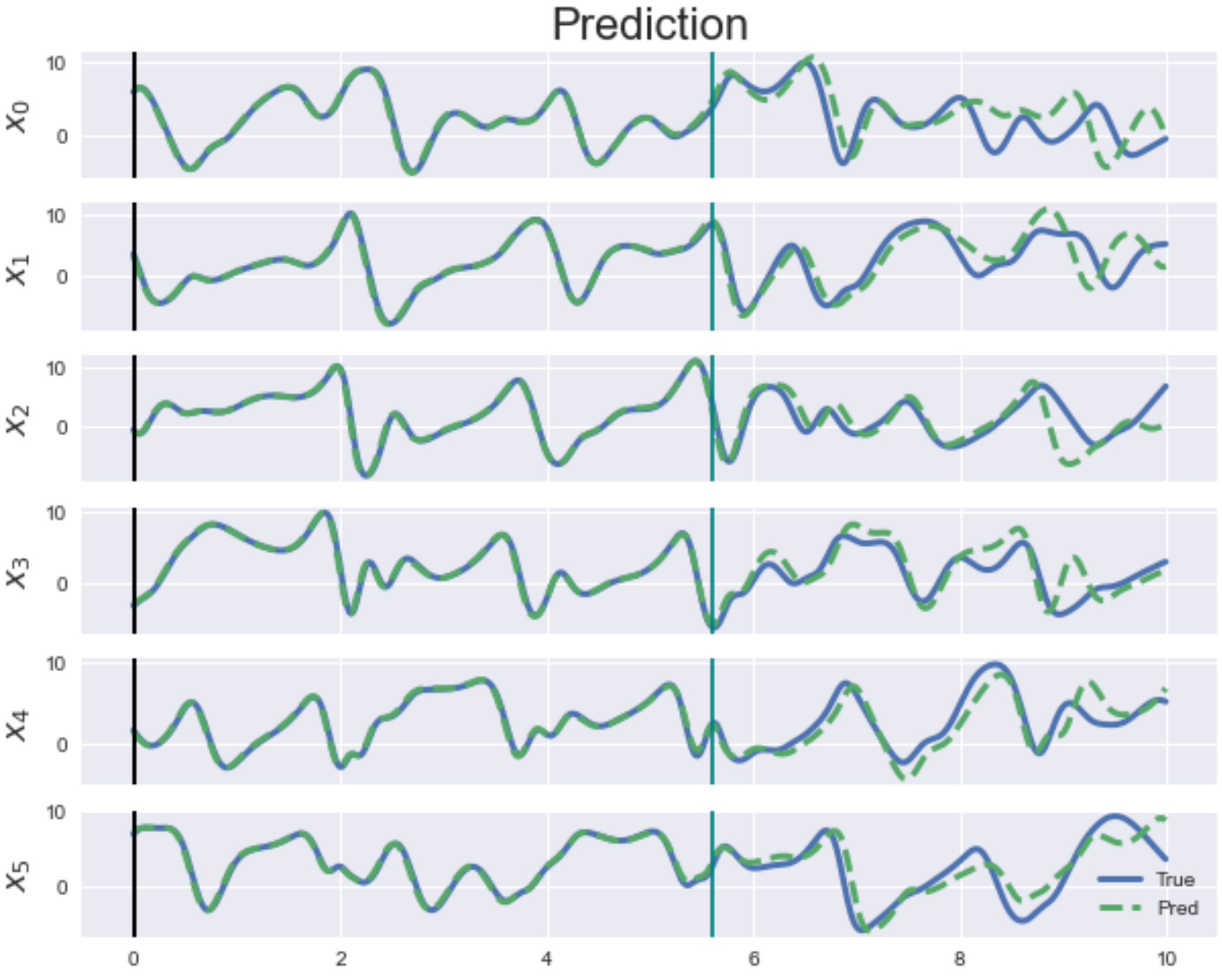}
    \caption{\label{fig:RNNtest}A free forecast of the RNN model 1 initialized from a random point in the test dataset, compared to the true trajectory. The valid prediction time (VPT) using $\epsilon$=0.2 is marked as a vertical line.}
\end{figure}

\begin{figure}
    \centering
    \includegraphics[width=0.8\textwidth]{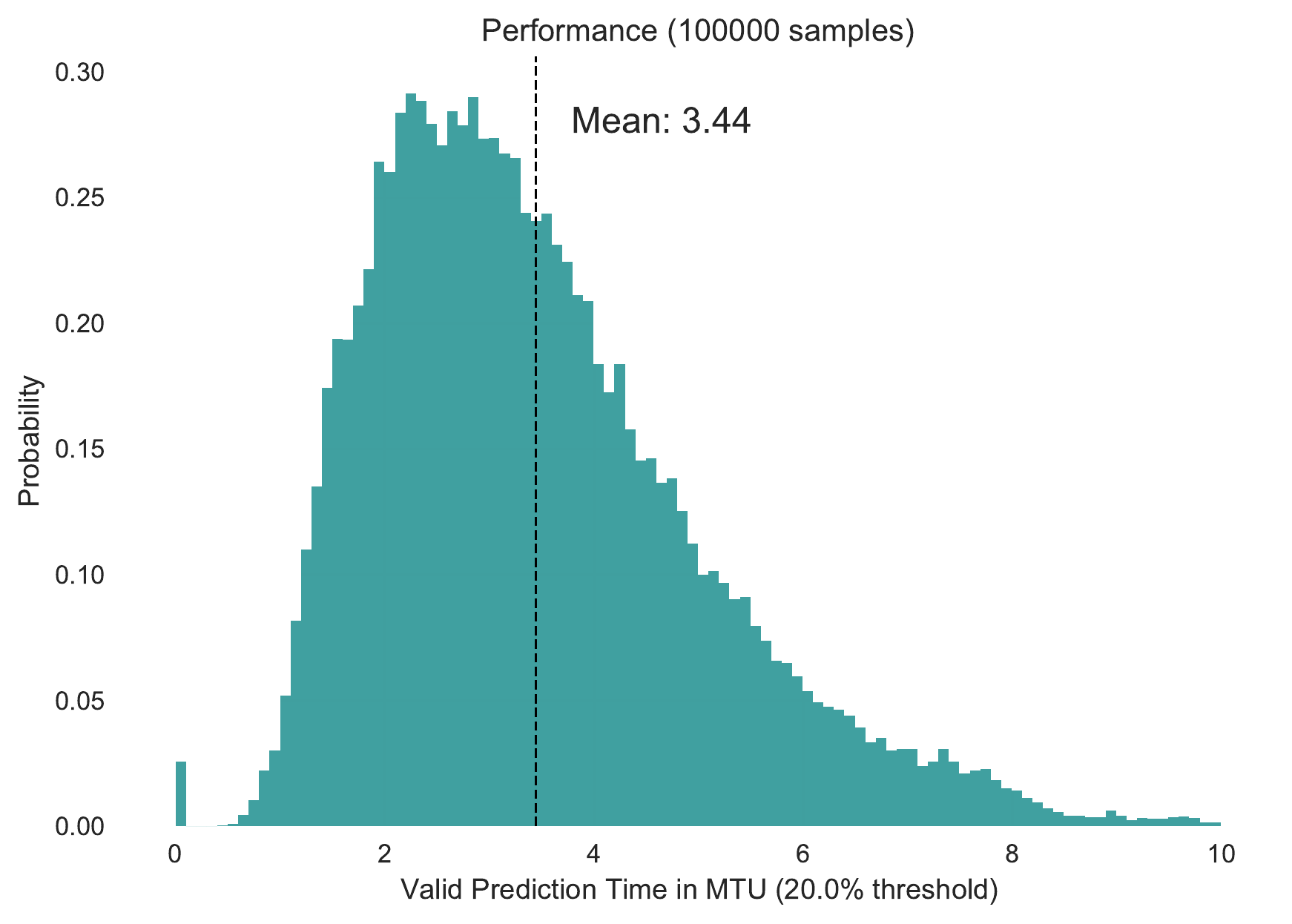}
    \caption{Histogram of valid prediction time (VPT) for RNN model 1 forecasts drawn from 100,000 initial conditions in the test dataset.}
    \label{fig:hist1}
\end{figure}

\begin{figure}
    \centering
    \includegraphics[width=0.8\textwidth]{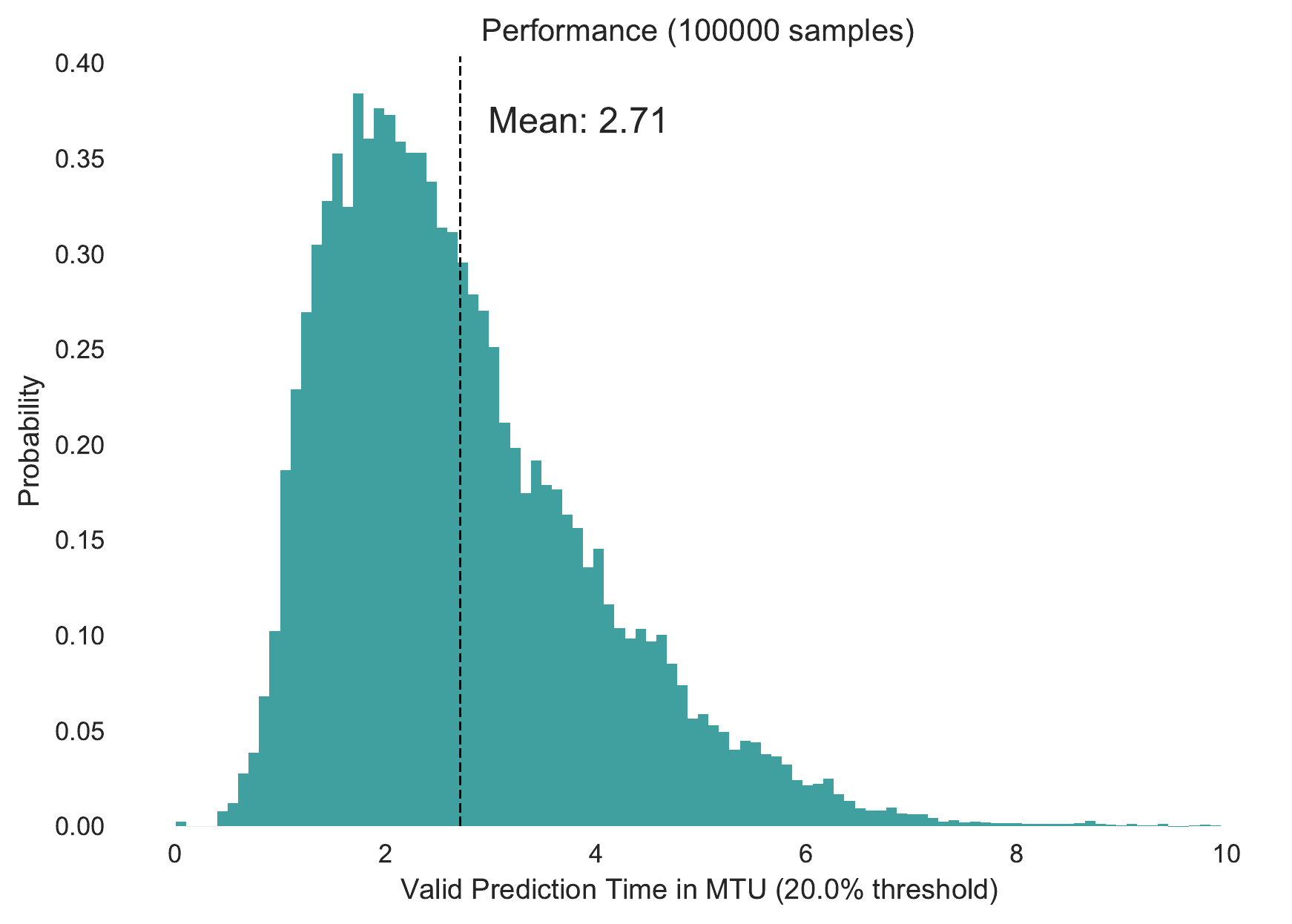}
    \caption{As in Figure \ref{fig:hist1}, but using RNN model 2. Note the model is skillful, but the mean VPT is reduced compared to model 1.}
    \label{fig:hist2}
\end{figure}

\begin{figure}
    \centering
    \includegraphics[width=0.6\textwidth]{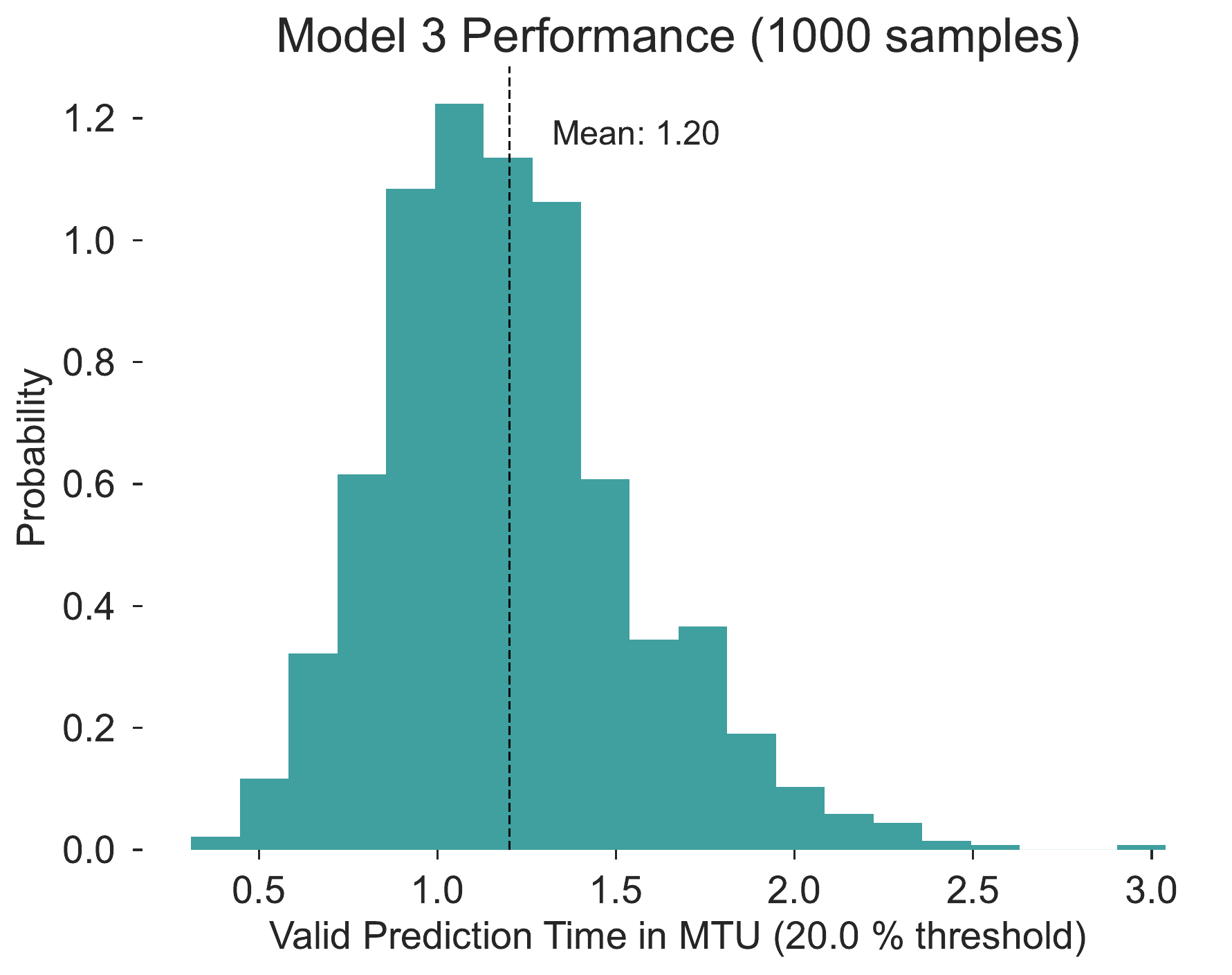}
    \caption{As in Figures \ref{fig:hist1} and \ref{fig:hist2}, but using RNN Model 3. Due to the computational cost of Model 3, only 1,000 sample forecasts are used. The model is still skillful, but the mean VPT is reduced when compared to Models 1 and 2.
    }
    \label{fig:hist3}
\end{figure}

We establish a `Valid Prediction Time' (VPT) as the length of time that the RMSE of any forecast starting at time $t_0$ remains below a threshold $\epsilon$. The VPT is defined precisely as,

\begin{equation}
    \sigma_i^{clim} = \sqrt{\sum_{t=t^{train}_0}^{t^{train}_N}{x_i(t)}},
\end{equation}
\begin{equation}
    x^f_{RMSE}(t,t_0) = \sqrt{\sum_{i=1}^{D}{\left(\frac{x_i^f(t,t_0) - x_i(t)}{\sigma_i^{clim}}\right)^2}},
\end{equation}
\begin{equation}
    VPT(t_0) = \max\{t : x^f_{RMSE}(t,t_0) < \epsilon, \, \forall t > t_0\}.
\end{equation}

where $\sigma_{i}^{clim}$ is the standard deviation in time of the $i^{th}$ model variable, $x_i^f(t,t_0)$ is the forecast from $t_0$ to $t$, $x_i$ is the true state, $D$ is the number of model variables, and $x_{RMSE}^f(t,t_0)$ is the corresponding RMSE of the forecast error at time $t$.
An example RNN forecast using RNN model 1 is shown in Figure \ref{fig:RNNtest}, with the VPT shown using $\epsilon = 0.2$. We reiterate that the Bayesian optimization is used to identify parameters that produce the best average forecast skill across the training dataset. Data assimilation experiments are applied using a separately generated test dataset. Figures \ref{fig:hist1} and \ref{fig:hist2} demonstrate the distribution of prediction skill for Models 1 and 2 initialized from 100,000 initial conditions from the test dataset. The RMSE is normalized by climatological variability, defined as the standard deviation in time of each model variable calculated over the training dataset.

%\newpage
\FloatBarrier

\bibliographystyle{unsrtnat}
\bibliography{ref.bib}  %%% Uncomment this line and comment out the ``thebibliography'' section below to use the external .bib file (using bibtex) .





% \end{thebibliography}

\end{document}